\documentclass[sigconf]{acmart}
\AtBeginDocument{%
  \providecommand\BibTeX{{%
    \normalfont B\kern-0.5em{\scshape i\kern-0.25em b}\kern-0.8em\TeX}}}

\copyrightyear{2025}
\acmYear{2025}
\setcopyright{acmlicensed}
\acmConference[KDD '25]{Proceedings of the 31st ACM SIGKDD Conference on Knowledge Discovery and Data Mining V.2}{August 3--7, 2025}{Toronto, ON, Canada}
\acmBooktitle{Proceedings of the 31st ACM SIGKDD Conference on Knowledge Discovery and Data Mining V.2 (KDD '25), August 3--7, 2025, Toronto, ON, Canada}
\acmDOI{10.1145/3711896.3737110}
\acmISBN{979-8-4007-1454-2/2025/08}

\usepackage{hyperref}
\hypersetup{
    colorlinks=true,
    linkcolor=black,
    urlcolor=black,
    citecolor=black
}

\def\equationautorefname~#1\null{Equation~(#1)\null}
\usepackage{url}
\usepackage{bm}
\usepackage{bbm}
\usepackage{xspace}
\usepackage{algorithmic}
\usepackage{algorithm}
\usepackage{mdwlist}    
\usepackage{paralist} 

\usepackage{adjustbox}
\usepackage{array}
\usepackage{booktabs}
\usepackage{multirow}

\usepackage{array,graphicx}
\usepackage{booktabs}
\usepackage{pifont}
\usepackage{color}

\usepackage{colortbl}
\definecolor{lightgray}{gray}{0.85}
\usepackage{multirow}

\usepackage{fancybox} 
\usepackage{amsmath}
\usepackage{amsfonts}

\usepackage{enumitem}

\newcommand{\alg}[1]{Alg.~#1}
\newcommand{\step}[1]{Line~#1}
\newcommand{\eq}[1]{Eq.~(#1)}
\newcommand{\tabl}[1]{Table~#1}
\newcommand{\fig}[1]{Fig.~#1}
\newcommand{\secton}[1]{Section~#1}
\newcommand{\apdx}[1]{Appendix~#1}

\newcommand{\myparaitemize}[1]{\noindent{\textbf{#1.}}}

\newcommand{\romanone}{$\mathrm{\hspace{.18em}i\hspace{.18em}}$\xspace}
\newcommand{\romantwo}{$\mathrm{\hspace{.08em}ii\hspace{.08em}}$\xspace}
\newcommand{\romanthree}{$\mathrm{i\hspace{-.08em}i\hspace{-.08em}i}$\xspace}

\newcommand{\hide}[1]{}

\newcommand{\method}{\textsc{RedLamp}\xspace}

\newcommand{\mts}{\bm{X}} 
\newcommand{\mtsreconst}{\bm{\hat{X}}} 
\newcommand{\mtsvector}{\bm{x}} 
\newcommand{\labl}{\bm{y}} 
\newcommand{\predlabl}{\bm{\hat{y}}} 
\newcommand{\clabl}{\bm{\tilde{y}}} 
\newcommand{\mask}{\bm{W}} 

\newcommand{\mtst}[1]{\mts^{(#1)}}
\newcommand{\mtsreconstt}[1]{\mtsreconst^{(#1)}}

\newcommand{\predlablt}[1]{\predlabl^{(#1)}}

\newcommand{\mtstk}[2]{\mts^{(#1,#2)}}
\newcommand{\mtsreconsttk}[2]{\mtsreconst^{(#1,#2)}}
\newcommand{\labltk}[2]{\labl^{(#1,#2)}}
\newcommand{\predlabltk}[2]{\predlabl^{(#1,#2)}}
\newcommand{\clabltk}[2]{\clabl^{(#1,#2)}}
\newcommand{\masktk}[2]{\mask^{(#1,#2)}}

\newcommand{\trainingset}{\mathcal{X}^{train}} 
\newcommand{\testset}{\mathcal{X}^{test}} 

\newcommand{\augset}{\mathcal{X}^{aug}} 
\newcommand{\auglabelset}{\mathcal{Y}^{aug}} 
\newcommand{\augmaskset}{\mathcal{W}^{aug}} 
\newcommand{\augfunc}{f} 
\newcommand{\augfunck}[1]{\augfunc^{(#1)}} 
\newcommand{\selectedvariables}{\textbf{dims}}
\newcommand{\selectedvariable}{dim}

\newcommand{\anomalyscore}{\bm{s}} 

\newcommand{\totallength}{T} 
\newcommand{\timestep}{L} 
\newcommand{\ndim}{D} 
\newcommand{\nclass}{K} 

\newcommand{\probnorm}{\alpha} 
\newcommand{\probanom}{\beta} 
\newcommand{\lossweight}{\gamma} 
\newcommand{\threshold}{\delta} 

\newcommand{\lossfunc}{\mathcal{L}} 
\newcommand{\lossfuncce}{\lossfunc_{CE}} 
\newcommand{\lossfuncmse}{\lossfunc_{MSE}} 

\newcommand{\idmat}{\bm{I}} 

\newcommand{\realnumber}{\mathbb{R}} 

\newcommand{\hadamard}{\circ} 

\newcommand{\minmax}{\mathop{\rm MinMax}\limits}

\newcommand{\ucr}{UCR\xspace}
\newcommand{\iops}{AIOps\xspace}
\newcommand{\smd}{SMD\xspace}
\newcommand{\smap}{SMAP\xspace}
\newcommand{\msl}{MSL\xspace}

\newcommand{\at}{AT\xspace}
\newcommand{\beatgan}{BeatGAN\xspace}
\newcommand{\isf}{IsolationForest\xspace}
\newcommand{\ocsvm}{OCSVM\xspace}
\newcommand{\lstmvae}{LSTM-VAE\xspace}
\newcommand{\tranad}{TranAD\xspace}
\newcommand{\usad}{USAD\xspace}
\newcommand{\imdiffusion}{IMDiffusion\xspace}
\newcommand{\dddr}{D3R\xspace}
\newcommand{\ncad}{NCAD\xspace}
\newcommand{\capa}{CutAddPaste\xspace}
\newcommand{\couta}{COUTA\xspace}

\begin{document}

\title{
Robust and Explainable Detector of Time Series Anomaly \\ via Augmenting Multiclass Pseudo-Anomalies
}


\author{Kohei Obata}
\email{obata88@sanken.osaka-u.ac.jp}
\affiliation{%
  \institution{SANKEN, University of Osaka}
  \state{Osaka}
  \country{Japan}
}

\author{Yasuko Matsubara}
\email{yasuko@sanken.osaka-u.ac.jp}
\affiliation{%
  \institution{SANKEN, University of Osaka}
  \state{Osaka}
  \country{Japan}
}

\author{Yasushi Sakurai}
\email{yasushi@sanken.osaka-u.ac.jp}
\affiliation{%
  \institution{SANKEN, University of Osaka}
  \state{Osaka}
  \country{Japan}
}

\renewcommand{\shortauthors}{Kohei Obata, Yasuko Matsubara, and Yasushi Sakurai}

\begin{abstract}
    Unsupervised anomaly detection in time series has been a pivotal research area for decades.
Current mainstream approaches focus on learning normality, on the assumption that all or most of the samples in the training set are normal.
However, anomalies in the training set (i.e., anomaly contamination) can be misleading.
Recent studies employ data augmentation to generate pseudo-anomalies and learn the boundary separating the training samples from the augmented samples.
Although this approach mitigates anomaly contamination if augmented samples mimic unseen real anomalies, it suffers from several limitations.
(1) Covering a wide range of time series anomalies is challenging.
(2) It disregards augmented samples that resemble normal samples (i.e., false anomalies).
(3) It places too much trust in the labels of training and augmented samples.
In response, we propose \method, which employs diverse data augmentations to generate multiclass pseudo-anomalies and learns the multiclass boundary.
Such multiclass pseudo-anomalies cover a wide variety of time series anomalies.
We conduct multiclass classification using soft labels, which prevents the model from being overconfident and ensures its robustness against contaminated/false anomalies.
The learned latent space is inherently explainable as it is trained to separate pseudo-anomalies into multiclasses.
Extensive experiments demonstrate the effectiveness of \method in anomaly detection and its robustness against anomaly contamination.

\end{abstract}



\maketitle

\section{Introduction}
    \label{010intro}
    \begin{figure*}[t]
    \centering
    \includegraphics[height=10em]{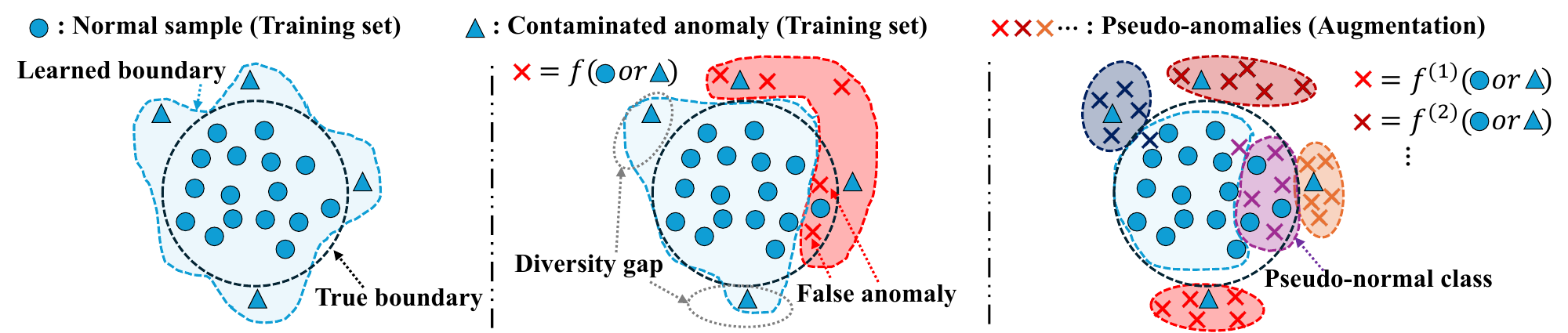} \\ 
    \hspace{1em} (a) Normality assumption \hspace{2em} (b) Binary anomaly assumption \hspace{2em} (c) Multiclass anomaly assumption \\ 
    \vspace{-1.em}
    \caption{
    Comparison of three assumptions.
    The true boundary represents the border between normal and anomaly samples in the test set that we want to predict.
    (a) Contaminated anomalies affect the learned boundary.
    (b) It generates pseudo-anomalies by a single data augmentation.
    The diversity gap and false anomalies degrade detection performance.
    (c) It employs diverse data augmentations and fills the diversity gap.
    Some data augmentation (purple) may be useless and form a pseudo-normal class.
    }
    \label{fig:problem}
    \vspace{-1.em}
\end{figure*}

Time series anomaly detection (TSAD) is an important field in data mining and analytics.
Its goal is to detect unexpected patterns or values within sequences that are significantly dissimilar to the majority.
It has numerous real-world applications, including but not limited to robot-assisted systems~\cite{lstmvae,robot}, engine monitoring~\cite{engine_lstmencdec}, and cyber-physical system maintenance~\cite{hs-tree}.
Anomalies are rare by nature, and labeling them from large quantities of data would be difficult and costly.
Therefore, TSAD is often formulated as an unsupervised problem with an unlabeled training set.

Unsupervised learning methods for TSAD, such as reconstruction-based methods~\cite{omnianomaly,anomalytransformer} and one-class classification~\cite{ocsvm}, typically rely on the \textit{normality assumption}, where the dataset is presumed to consist of all or mostly normal samples.
However, the presence of anomalies in the training set (i.e., anomaly contamination) is inevitable~\cite{mitigatecontamination,couta} and is not sufficiently addressed under this assumption.
Consequently, as illustrated in \fig{\ref{fig:problem}} (a),  normality assumption-based methods are vulnerable to anomaly contamination and easily lead to overgeneralization (i.e., high false negative rate).

Recent studies, particularly in computer vision~\cite{oe,cutpaste,dream}, have utilized data augmentation for anomaly detection.
This approach is based on the \textit{binary anomaly assumption}, which creates pseudo-anomalies through data augmentation, typically through manually designed functions such as cutout and jittering, and learns the boundary separating given training samples (i.e., normal) and augmented samples (i.e., anomaly).
Recently, \capa~\cite{cutaddpaste} employed time series-specific anomaly knowledge to design data augmentation for TSAD.
Although existing binary anomaly assumption-based methods can mitigate the negative impact of anomaly contamination if pseudo-anomalies resemble contaminated anomalies, they face the following three issues:
\begin{enumerate}[left=0pt]
    \item \textit{Diversity gap}:
    It is challenging to mimic real anomalies, as they vary across datasets and are unknown in advance~\cite{hyperparameteraug}.
    Thus, pseudo-anomalies often fail to encompass all anomalies, leaving an uncovered space we call the diversity gap.
    As shown in \fig{\ref{fig:problem}} (b), the diversity gap makes the model susceptible to contaminated anomalies, leading to a high false negative rate.
    \item \textit{False anomaly}:
    Data augmentation often generates pseudo-anomalies that should be considered normal, which we refer to as false anomalies~\cite{pseudoanomaly}.
    As shown in \fig{\ref{fig:problem}} (b), the model overfits false anomalies and leads to a high false positive rate.
    \item \textit{Overconfidence on the label}:
    Existing methods overly trust the labels of both given samples and pseudo-anomalies; namely, they assign them with binary hard labels.
    These methods fail to account for the contaminated/false anomalies, leading to overconfidence prediction and an inaccurate decision boundary.
\end{enumerate}

Thus, in this paper, we propose a \textit{\textbf{R}obust and \textbf{E}xplainable \textbf{D}etector of time series anoma\textbf{L}y via \textbf{A}ugmenting \textbf{M}ulticlass \textbf{P}seudo-anomalies} called \method
\footnote{The code is available online: \url{https://github.com/KoheiObata/RedLamp}}
to address the above issues.
The proposed method is based on the \textit{multiclass anomaly assumption}, which creates various types of pseudo-anomalies through multiple data augmentations and learns a multiclass decision boundary, as shown in \fig{\ref{fig:problem}} (c).
We prevent the diversity gap from occurring by utilizing diverse data augmentations tailored for time series (e.g., spike, speedup, etc.). 
To address contaminated/false anomalies, we treat them as label noise and employ soft labels for multiclass classification.
The soft labels allow the model to consider whether a potential anomaly sample in the training set is mislabeled as a normal sample (i.e., anomaly contamination) or vice versa (i.e., false anomaly), thus preventing the model from being overconfident but ensuring robustness against contaminated/false anomalies.
Our model predicts reconstruction in addition to multiclass labels from the common embeddings to utilize the strong detection capability of the reconstruction-based approach.
Therefore, we train the model with reconstruction loss and cross-entropy loss.
Unlike binary anomaly assumption-based methods, the learned latent space of \method is inherently explainable as it is trained to separate pseudo-anomalies into multiclasses.
Finally, we introduce the anomaly score, which consists of the reconstruction error and the adjusted anomaly-class score.
Depending on the data, data augmentation may not be effective (i.e., it generates only false anomalies), as in \fig{\ref{fig:problem}} (c).
We refer to the anomaly class containing many normal samples as the \textit{pseudo-normal class}.
Our adjusted anomaly-class score addresses the issue by considering frequently detected classes as normal.

In summary, this paper makes the following contributions:
\begin{itemize}[left=0pt]
    \item \textbf{Multiclass anomaly assumption}: 
    We highlight the \textit{diversity gap}, \textit{false anomalies}, and \textit{overconfidence} issues associated with the existing binary anomaly assumption-based methods and demonstrate how \method addresses them.
    We are the first to solve TSAD by using multiclass classification.
    \item \textbf{Effectiveness}: Extensive experiments on five real-world datasets shows that \method outperforms baseline approaches by $17\%$ in terms of VUS-PR.
    \item \textbf{Robustness}: 
    With $5\%$ contaminated anomalies added to the UCR dataset, \method exhibits only a $1.8\%$ performance drop in VUS-PR, whereas \capa incurs a $15.5\%$ drop.
    \item \textbf{Explainability}:
    \method provides an explainable latent space that gives us insight into which anomaly class the real anomalies are close to through its visualization.
\end{itemize}

\section{Related Work}
    \label{020related}
    
\subsection{Anomaly Detection in Time Series}
TSAD has been extensively studied over several decades~\cite{survey_acm,survey_vldb,survey_units}.
Initial machine learning methods include \isf~\cite{isolationforest} and \ocsvm~\cite{ocsvm}. 
Subsequently, many deep anomaly detection models have been proposed for capturing anomalous behavior in an unsupervised manner.
Most are based on the normality assumption that considers the majority of the samples in the training set to be normal. 
As our model is partially based on reconstruction, we focus on reconstruction-based methods, which aim to learn the latent representation for the entire time series that can reconstruct normal samples while failing with anomaly samples.
Deep generative techniques, including autoencoders (AE)~\cite{usad,asynchronous,interfusion}, GAN~\cite{tanogan,beatgan}, and diffusion models~\cite{diffad,dddr}, are often used to build reconstruction-based methods.
For example, LSTM was incorporated into VAE to learn temporal features~\cite{lstmvae,omnianomaly}.
\imdiffusion~\cite{imdiffusion} integrates the imputation method with a grating masking strategy and employs diffusion models to generate masked sequences.
In addition to generative techniques, transformer architecture~\cite{anomalytransformer,dcdetector,tranad} and graph relational learning~\cite{gdn,fusagnet} have also been explored.
These methods control the trade-off between concise representation (i.e., regularization) and representation power (i.e., reconstruction).
Although they provide good results for multivariate time series benchmark datasets~\cite{elephant}, controlling such trade-offs remains challenging.
As a result, they tend to miss pattern-wise anomalies, especially when the differences from a normal sample are subtle.
Moreover, they suffer from overgeneralization under anomaly contamination as they are based on the normality assumption.

\subsection{Data Augmentation for Anomaly Detection}
Data augmentation increases the diversity of the training set and improves the model’s ability to generalize to unseen data~\cite{augsurvey,augempirical}.
Generally, it assumes that the augmented samples cover unexplored input space while maintaining labels.
For example, \beatgan~\cite{beatgan} is a reconstruction-based method proposed for detecting anomalies in ECG beats that uses time series warping for data augmentation and augmented samples as normal samples.

In anomaly detection, data augmentation is also used to handle data imbalance; that is, pseudo-anomalies are generated to replace an unsupervised task with a supervised binary classification task.
This approach relies on a binary anomaly assumption that learns the boundary that distinguishes between the training samples and the augmented samples.
The more the pseudo-anomalies mimic the nature of the true anomalies in the test set, the better the detection performance becomes.
Many augmentation techniques are proposed in computer vision~\cite{dream}, such as cutout~\cite{cutout} and outlier exposure (OE)~\cite{oe}.
Li et al.~\cite{cutpaste} introduce CutPaste and CutPaste-Scar as data augmentations and demonstrate that conducting a 3-class classification using both augmentations performs the best.
However, these augmentations are tailored to the image and cannot be directly applied to time series.

The aim of several studies has been to generate time series-specific augmentations~\cite{anomalybert,coca}.
For example, \ncad~\cite{ncad} designs data augmentation for TSAD utilizing OE, point outlier, and window mixup and then applies the hypersphere classifier.
Inspired by CutPaste, Wang et al.~\cite{cutaddpaste} propose \capa, which is designed to generate diverse anomalies by mimicking five different types of anomalies (e.g., global point, seasonal) based on the prior taxonomy~\cite{tsadtaxonomy} in a single data augmentation.
Unlike reconstruction-based methods, \capa can detect pattern-wise anomalies if suitable pseudo-anomalies are generated.
However, \capa requires many hyperparameters to be set to generate adequate augmentation for each dataset~\cite{hyperparameteraug} and is affected by anomaly contamination when there is a diversity gap. 
To tackle the anomaly contamination issue, \couta~\cite{couta} employs data augmentation to create naive anomaly samples and penalizes irregular samples in the training set.
However, these binary anomaly assumption-based methods are overly confident as regards the generated pseudo-anomaly and ignore frequently occurring false anomalies, leading to degraded detection accuracy.

\begin{table}[t]
    \caption{
    Capabilities of \method and anomaly detection methods that employ data augmentation.
    Nor./Ano. generates normal/anomaly samples through data augmentation. 
    }
    \vspace{-1em}
    \label{table:req}
    \centering
    \resizebox{1.0\linewidth}{!}{
        \begin{tabular}{c|ccccccc}
        \toprule
            & \rotatebox{90}{OE~\cite{oe}} & \rotatebox{90}{CutPaste~\cite{cutpaste}} & \rotatebox{90}{BeatGAN~\cite{beatgan}} & \rotatebox{90}{NCAD~\cite{ncad}} & \rotatebox{90}{CutAddPaste~\cite{cutaddpaste}} & \rotatebox{90}{COUTA~\cite{couta}} & \rotatebox{90}{\textbf{\method}} \\
            \midrule
            Domain & CV & CV & TS & TS & TS & TS & TS\\
            \rowcolor[gray]{0.8}
            Data augmentation & Ano. & Ano. & Nor. & Ano. & Ano. &  Ano. & Ano. \\
            Multiclass anomalies & - & 3-way & - & - & - & - & \checkmark\\
            \rowcolor[gray]{0.8}
            Anomaly diversity & - & - & - & - & \checkmark & - & \checkmark\\
            Anomaly contamination & - & - & - & - & - & \checkmark & \checkmark\\
            \rowcolor[gray]{0.8}
            False anomaly & - & - & - & - & - & - & \checkmark\\
            \bottomrule
        \end{tabular}
    }
    \vspace{-1em}
\end{table}
\myparaitemize{Significance of this work}
\tabl{\ref{table:req}} shows the relative advantages of \method compared with existing methods employing data augmentation.
\method is the only method that exploits multiple time series-specific data augmentations and learns to classify them.
Therefore, generated pseudo-anomalies cover a variety of time series-specific anomalies, and the learned latent space is inherently explainable.
\method is robust to anomaly contamination and constitutes the first approach to focus on a false anomaly.

\section{Proposed \method}
    \label{040model}
    
\subsection{Problem Definition}
In this paper, we focus on unsupervised anomaly detection for time series.
We adopt a local contextual window to model their temporal dependency, which is the common practice with most deep TSAD methods.
Consider a multivariate time series of $\ndim$ features collected over $\totallength$ timestep,
$\mts = \{ \mtsvector_1,\mtsvector_2,\dots,\mtsvector_\totallength \} \in \realnumber^{\ndim \times \totallength}$.
We define the training set $\trainingset = \{ \mtst{\timestep},\mtst{\timestep+1},\dots,\mtst{\totallength} \}$,
where $\mtst{t} = \{ \mtsvector_{t-\timestep+1},\dots,\mtsvector_t \} \in \realnumber^{\ndim \times \timestep}$ represents a $t$-th instance segmented by a sliding window of size $\timestep$.
For simplicity, we consider that the stride is set at 1.
$\mtst{t}_{i,j}$ indicates the element at the $i$-th feature and $j$-th timestep.
The goal of TDAD is to identify anomalous timesteps in a test set $\testset$ by training the model on an unlabeled training set.
The test set is also split into sub-sequences using the same window size $\timestep$.
To detect the anomaly, the model aims to predict the anomaly scores $\anomalyscore(t)$ of the instance $\mtst{t} \in \testset$, where higher scores are more anomalous.

\begin{figure*}[t]
    \centering
    \begin{tabular}{cc} 
        \hspace{-2em}
        \includegraphics[height=15em]{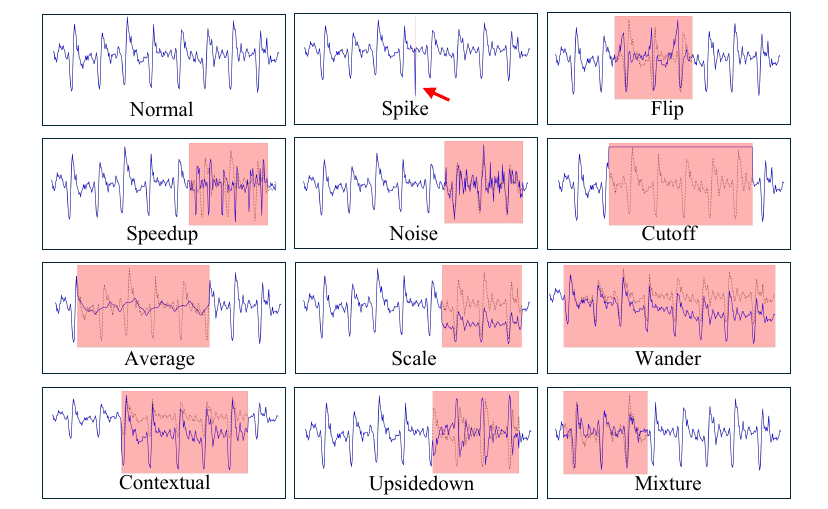} & 
        \hspace{-1em}
        \includegraphics[height=15em]{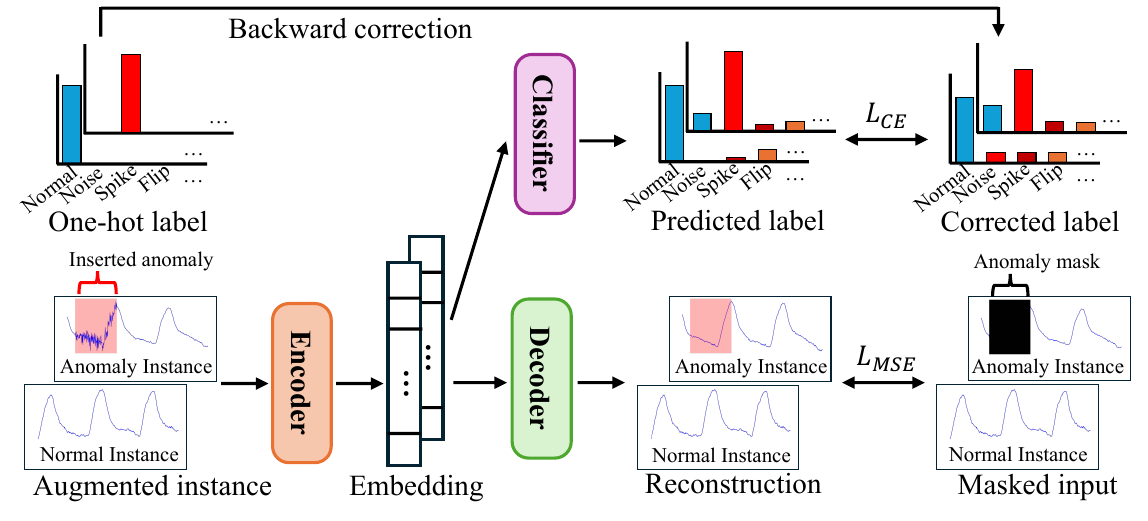} \\ 
        (a) Example of augmented instances & (b) Model architecture \\ 
    \end{tabular}
    \vspace{-1.em}
    \caption{
    (a) Augmentation includes one normal and 11 different types of anomalies. 
    A gray dotted line indicates a sequence before augmentation, and a red background indicates the range of the inserted anomaly.
    (b) \method trains the model with the augmented training set consisting of augmented instances, one-hot labels, and anomaly masks.
    The model predicts anomaly-free reconstruction and multiclass labels from the same embedding.
    }
    \label{fig:overview}
    \vspace{-1.em}
\end{figure*}

\subsection{Overall Framework} \label{sec:overview}
We propose \method as the first multiclass anomaly assumption-based method for TSAD.
\method is robust against anomaly contamination and overcomes the challenges that face binary anomaly assumption-based methods, which are the diversity gap, false anomaly, and overconfidence in labels.
The core idea is to generate multiclass pseudo-anomalies through diverse data augmentations, enabling the model to learn a multiclass decision boundary.

An overview of the \method framework is shown in \fig{\ref{fig:overview}}.
The proposed method begins by augmenting the training set to create an augmented instance set that includes normal instances and 11 types of pseudo-anomaly instances, as depicted in \fig{\ref{fig:overview}} (a).
Each augmented instance is accompanied by a one-hot label and an anomaly mask that indicates the location of the inserted anomaly (\secton{\ref{sec:augmentation}}). 
Using the augmented training set, we train a model comprising a CNN-based encoder and decoder and a 2-layered MLP classifier.
Details of the model architecture can be found in \apdx{\ref{apdx:implementation}}.
As shown in \fig{\ref{fig:overview}} (b), the augmented instances are first projected into embeddings by the encoder.
The decoder then generates reconstructions while the classifier predicts labels.
The decoder learns to reconstruct only the normal parts of the instances by utilizing the anomaly masks so that we can utilize the strong detection capability of a reconstruction-based approach.
To tackle the challenges of contaminated/false anomalies, we introduce backward correction and transform one-hot labels into corrected soft labels, which the classifier learns to predict (\secton{\ref{sec:labelsmothing}}).
Therefore, model parameters are learned via the loss function composed of masked reconstruction loss and cross-entropy loss (\secton{\ref{sec:training}}).
Our anomaly score consists of reconstruction error and adjusted anomaly-class score that addresses the pseudo-normal class by considering a frequently observed class as normal (\secton{\ref{sec:score}}).

\begin{algorithm}[t]
    \small
    \caption{\textsc{Data Augmentation}}
    \label{alg:augmentation}
    \begin{algorithmic}[1]
        \STATE {\bf Input:}
        Training set $\trainingset = \{ \mtst{\timestep},\mtst{\timestep+1},\dots,\mtst{\totallength} \}$
        \STATE {\bf Output:}
        Augmented instance set $\augset = \{ \mtstk{\timestep}{1}, \dots, \mtstk{\totallength}{\nclass} \}$ \\
        \hspace{3.4em} Label set $\auglabelset = \{ \labltk{\timestep}{1}, \dots, \labltk{\totallength}{\nclass} \}$ \\
        \hspace{3.4em} Anomaly mask set $\augmaskset = \{ \masktk{\timestep}{1}, \dots, \masktk{\totallength}{\nclass} \}$
        \STATE Initialize $\augset, \auglabelset, \augmaskset = \phi,\phi,\phi$;
        \FOR{$t=\timestep:\totallength$}
            \FOR{$k=1:\nclass$}
                \STATE 
                $\labltk{t}{k} = \mathbf{0}$;\; $\labltk{t}{k}_{k} = 1$;\; $\masktk{t}{k} = \mathbf{0}$;
                \STATE 
                $\selectedvariables = Uni( set=\{1,\dots,\ndim\}, size=Uni(\{1,\dots,\ndim\}, 1))$; \label{line:selectvariable}
                \FOR{$\selectedvariable$ in $\selectedvariables$}
                    \STATE 
                    $st, ed = Sort( Uni(\{1,\dots,\timestep\}, 2) )$; \label{line:selectrange}
                    \STATE 
                    $\mtstk{t}{k} = \augfunck{k}(\mtst{t}, \selectedvariable, st, ed)$; \label{line:augment}
                    \STATE \textbf{if} $\augfunck{k}$ is Spike \textbf{then}
                    $\masktk{t}{k}_{\selectedvariable, st} = 1$;
                    \STATE \textbf{else if} $\augfunck{k}$ is Wander \textbf{then}
                    $\masktk{t}{k}_{\selectedvariable, st:} = \mathbf{1}$;
                    \STATE \textbf{else}
                    $\masktk{t}{k}_{\selectedvariable, st:ed} = \mathbf{1}$; \label{line:mask}
                    \STATE \textbf{end if}
                \ENDFOR
                \STATE Append $\mtstk{t}{k}, \labltk{t}{k}, \masktk{t}{k}$
                to $\augset, \auglabelset, \augmaskset$;
            \ENDFOR
        \ENDFOR
        \RETURN $\augset, \auglabelset, \augmaskset$;
    \end{algorithmic}
    \normalsize
\end{algorithm}

\subsection{Time Series Anomaly Data Augmentation} \label{sec:augmentation}

Filling the diversity gap is important in terms of tackling anomaly contamination, and it also enhances detection performance (we discuss in \secton{\ref{sec:q3} and \ref{sec:q4}}).
Instead of trying to mimic the unseen real anomalies, we aim to cover a wide range of time series anomalies.
We begin by introducing the algorithm and then elaborate on the data augmentations for time series-specific anomalies.

\subsubsection{Algorithm}
We generate diverse pseudo-anomalies using $\nclass$ different data augmentations $\{\augfunck{1}, \dots, \augfunck{\nclass}\}$ tailored for time series, where $\nclass$ is the number of classes, which consists of one normal and $\nclass-1$ anomaly classes.
The data augmentation process inputs the training set $\trainingset$ and outputs the augmented training set,
which consists of augmented instance set $\augset = \{ \mtstk{\timestep}{1}, \dots, \mtstk{\totallength}{\nclass} \}$,
label set $\auglabelset = \{ \labltk{\timestep}{1}, \dots, \labltk{\totallength}{\nclass} \}$,
and anomaly mask set $\augmaskset = \{ \masktk{\timestep}{1}, \dots, \masktk{\totallength}{\nclass} \}$.
An augmented instance $\mtstk{t}{k}$ is generated by applying the $k$-th data augmentation $\augfunck{k}$ to the instance $\mtst{t}$, where $k=1$ corresponds to the normal augmentation.
$\labltk{t}{k} \in \{0,1\}^{|\nclass|}$ is the one-hot vector, with the $k$-th element set at $1$.
The anomaly mask $\masktk{t}{k} \in \realnumber^{\ndim \times \timestep}$ indicates the location of the inserted anomalies, where $1$ denotes an anomaly and $0$ denotes an anomaly-free region.

\alg{\ref{alg:augmentation}} shows the overall procedure for data augmentation.
For each training instance $\mtst{t}$, it begins by randomly selecting $1 \sim \ndim$ features onto which the function will be applied (\step{\ref{line:selectvariable}}).
For each selected feature $\selectedvariable$, an anomaly range (from $st$ to $ed$) is randomly determined (\step{\ref{line:selectrange}}), and finally, a data augmentation $\augfunck{k}$ is applied (\step{\ref{line:augment}}).
We apply each data augmentation once to an instance; thus, the size of each pseudo-anomaly class is in a 1:1 ratio relative to the training set size.

\subsubsection{Data augmentation}
Inspired by previous work~\cite{unsupervisedTSAD}, we design 12 data augmentations, as illustrated in \fig{\ref{fig:overview}} (a).
These data augmentations, which consider time series-specific anomaly knowledge, generate anomalies of different traits, including five common types of time series anomalies~\cite{tsadtaxonomy}.
We describe them below using Python-format equations:
\begin{enumerate}[left=0pt]
    \item Normal: Uses the original data without modifications ($k=1$).
    \\$\mtstk{t}{k}_{\selectedvariable} = \mtst{t}_{\selectedvariable}$ .
    \item Spike: Adds a sharp spike at a selected point.
    \\$\mtstk{t}{k}_{\selectedvariable,st} = \mtst{t}_{\selectedvariable,st} + a$, where $ a \sim \mathcal{N}(0,1)$ .
    \item Flip: Flips a selected range horizontally.
    \\$\mtstk{t}{k}_{\selectedvariable,st:ed} = \mtst{t}_{\selectedvariable,st:ed}[::-1]$ .
    \item Speedup: Changes the frequency of the data within a selected range by speeding it up (2x) or slowing it down (1/2x).
    \\$\mtstk{t}{k}_{\selectedvariable,st:ed} = \mtst{t}_{\selectedvariable,st:ed+length}[::2]$ if 2x ,
    \\$\mtstk{t}{k}_{\selectedvariable,st:ed} = Stretch(\mtst{t}_{\selectedvariable,st:ed-(length/2)}; [length/2] \rightarrow [length])$ if 1/2x .
    \item Noise: Adds random noise sampled from a normal distribution.
    \\$\mtstk{t}{k}_{\selectedvariable,st:ed} = \mtst{t}_{\selectedvariable,st:ed} + \mathbf{a}$, where $\mathbf{a} \sim \mathcal{N}(\mathbf{0},0.1\idmat)$ .
    \item Cutoff: Replaces data with a uniform value between the minimum and maximum values of a selected range.
    \\$\mtstk{t}{k}_{\selectedvariable,st:ed} = Uni( \{ min(\mtst{t}_{\selectedvariable,st:ed}),\dots,max(\mtst{t}_{\selectedvariable,st:ed})\}, 1 )$.
    \item Average: Applies a moving average of window size $w$.
    \\$\mtstk{t}{k}_{\selectedvariable,st:ed} = MovingAverage(\mtst{t}_{\selectedvariable,st:ed}, w=\timestep/5)$ .
    \item Scale: Scales the values by a randomly generated factor.
    \\$\mtstk{t}{k}_{\selectedvariable,st:ed} = a*\mtst{t}_{\selectedvariable,st:ed}$, where $a \sim \mathcal{N}(1,1)$ .
    \item Wander: Adds a linear trend and then shifts the baseline.
    \\$\mtstk{t}{k}_{\selectedvariable,st:ed} = \mtst{t}_{\selectedvariable,st:ed} + np.linspace(0, a, length)$,
    \\$\mtstk{t}{k}_{\selectedvariable,st:} += a$, where $a \sim \mathcal{N}(0,1)$ .
    \item Contextual: Scales a selected range and shifts the baseline.
    \\$\mtstk{t}{k}_{\selectedvariable,st:ed} = a*\mtst{t}_{\selectedvariable,st:ed} + b$, where $a \sim \mathcal{N}(1,1), b \sim \mathcal{N}(0,1)$.
    \item Upsidedown: Inverts a selected range. 
    \\$\mtstk{t}{k}_{\selectedvariable,st:ed} = 2*Mean(\mtst{t}_{\selectedvariable,st:ed}) - \mtst{t}_{\selectedvariable,st:ed}$ .
    \item Mixture: Swaps a selected range with another segment of the same length from a different timestep.
    \\$\mtstk{t}{k}_{\selectedvariable,st:ed} = \mtst{t'}_{\selectedvariable,st:ed}$ , where $t' \in \{\timestep,...,\totallength\}$ ,
\end{enumerate}
where $\selectedvariable$ is a selected feature, $st, ed$ are the starting and ending points of the anomaly, and $length = ed - st$ is the length of the inserted anomaly.

\subsection{Label Refurbishment} \label{sec:labelsmothing}
The definition of anomaly varies across datasets, making it difficult to design data augmentations that align with the real anomalies (see \apdx{\ref{apdx:augmentation}}).
Thus, pseudo-anomalies often include false anomalies, e.g., applying a short-range moving average to data with a longer periodicity may not be meaningful.
Additionally, anomaly contamination in the training set is inevitable.
These issues can be viewed as label noise;
that a potential anomaly instance is mislabeled as normal (i.e., contaminated anomaly) or a potential normal instance is mislabeled as pseudo-anomaly (i.e., false anomaly).

To address contaminated/false anomalies, we convert hard labels into soft labels by label refurbishment~\cite{refurbishment}, which is a technique often used to boost classification performance under label noise.
Since our pseudo-anomalies are generated from normal instances, we utilize backward correction~\cite{backward}, which can include the probability of misclassifying one class as another.
The corrected label is formulated as follows:
\begin{align}
    \clabltk{t}{k} &=
    \begin{cases}
        (1 - \probnorm - \nclass\probanom)\labltk{t}{k} + \probnorm + \probanom & \text{if $k = 1$}, \\
        (1 - \probnorm - \nclass\probanom)\labltk{t}{k} + \probanom & \text{otherwise},
    \end{cases} \label{eq:labelsmooth}
\end{align}
where $\probnorm$ represents the probability that a pseudo-anomaly is misclassified as normal,
and $\probanom$ denotes the probability that one class is misclassified as another.
The soft labels prevent the model from producing overconfident predictions, thus enhancing its robustness against false anomalies and anomaly contamination.

\subsection{Objective Function} \label{sec:training}
We describe the loss function \eq{\ref{eq:totalloss}}, comprises masked reconstruction loss \eq{\ref{eq:lossmse}} and cross-entropy loss \eq{\ref{eq:lossce}}, contributing to a rich latent space for separating pseudo-anomalies into multiclasses.

\subsubsection{Masked reconstruction loss} \label{sec:reconstruction}
The decoder predicts the reconstruction $\mtsreconsttk{t}{k}$ of augmented instance $\mtstk{t}{k}$ from the embedding.
The masked reconstruction loss is used to minimize the error in anomaly-free regions by removing inserted anomalies using an anomaly mask $\masktk{t}{k}$.
This loss allows the embeddings to capture the features of normal behavior in the latent space.
The masked reconstruction loss is defined as:
\begin{align}
    \lossfuncmse = || (\textbf{1}-\masktk{t}{k})\hadamard(\mtstk{t}{k} - \mtsreconsttk{t}{k}) ||^{2}_{2},  \label{eq:lossmse}
\end{align}
where $\hadamard$ denotes element-wise multiplication.

\subsubsection{Cross-entropy loss} \label{sec:classification}
The classifier predicts label $\predlabltk{t}{k}$ for $\mtstk{t}{k}$ from the embedding.
The cross-entropy loss is calculated by comparing the predicted label with the corrected label $\clabltk{i}{k}$.
This loss helps prevent the model from overfitting contaminated/false anomalies and encourages the embeddings to separate into distinct classes in the latent space. 
The cross-entropy loss is defined as:
\begin{align}
    \lossfuncce = -\sum_{i=1}^{\nclass} \big( \clabltk{t}{k}_{i} log(\predlabltk{t}{k}_{i}) \big).  \label{eq:lossce}
\end{align}

\subsubsection{Total loss} \label{sec:totalloss}
Our loss function is the weighted sum of the above two losses and is defined as:
\begin{align}
    \lossfunc= \lossweight\lossfuncce + (1-\lossweight)\lossfuncmse,  \label{eq:totalloss}
\end{align}
where $\lossweight$ is a hyperparameter that adjusts the weight of the cross-entropy loss.

\subsection{Anomaly Scoring} \label{sec:score}
Finally, we propose an anomaly scoring function \eq{\ref{eq:score}} that integrates reconstruction error \eq{\ref{eq:reconstscore}} and adjusted anomaly-class score \eq{\ref{eq:cescore}}, which enables the detection of various types of anomalies.
The score is expected to be large when anomalies are input.

\subsubsection{Reconstruction error} \label{sec:reconstructionerror}
As with a typical reconstruction-based approach, our model learns to reconstruct anomaly-free parts.
Thus, the model is expected to excel as regards reconstruction in normal parts but struggle in anomalous parts.
Reconstruction error is computed by comparing the input instance with its reconstruction:
\begin{align}
    \anomalyscore_{MSE}(t) &= || \mtst{t} - \mtsreconstt{t} ||^{2}_{2}. \label{eq:reconstscore}
\end{align}

\subsubsection{Adjusted anomaly-class score} \label{sec:adjustedscore}
Some data augmentations may not disrupt the normality of the instance (check \fig{\ref{fig:tsne}} (b) (i)).
To address such pseudo-normal classes, we consider anomaly classes that are frequently detected in the test set as normal.
We define frequent anomaly adjustment (FAA) in \eq{\ref{eq:faa}}.
After applying FAA to the predicted label, the adjusted anomaly-class score is calculated as the sum of anomaly-class scores:
\begin{align}
    FAA(\predlablt{t}_{i}) &= 
    \begin{cases}
        0 & \text{if $\frac{1}{\totallength-\timestep+1}\sum_{j=\timestep}^{\totallength} \predlablt{j}_{i} > \threshold$}, \\
        \predlablt{t}_{i} & \text{otherwise}, 
    \end{cases} \label{eq:faa} \\
    \anomalyscore_{CE}(t) &= \sum_{i=2}^{\nclass} FAA(\predlablt{t}_{i}), \label{eq:cescore}
\end{align}
where pseudo-anomalies detected above the threshold $\threshold$ across the entire test set are considered normal (we set $\threshold=0.05$ by default).

\subsubsection{Total score} \label{sec:totalscore}
Our anomaly score is calculated by adding the two scores together after applying min-max normalization to them:
\begin{align}
    \anomalyscore(t) &= \frac{1}{2}\minmax_{\totallength}(\anomalyscore_{MSE}(t)) + \frac{1}{2}\minmax_{\totallength}(\anomalyscore_{CE}(t)).  \label{eq:score}
\end{align}

\section{Experiments}
    \label{050experiments}
    
To demonstrate the efficacy of \method, five aspects are investigated empirically:
\\ \textbf{Q1. Effectiveness}: How does \method perform compared with state-of-the-art baselines in TSAD?
\\ \textbf{Q2. Ablation study}: Do the proposed components contribute to better detection performance?
\\ \textbf{Q3. Sensitivity}: How do hyperparameters, especially the number of data augmentations, influence the anomaly detection results?
\\ \textbf{Q4. Robustness}: How does the robustness of \method behave w.r.t. various anomaly contamination levels of the training set?
\\ \textbf{Q5. Explainability}: How well does the learned latent space help us understand the real anomalies?

\subsection{Experimental Settings} \label{sec:setting}

\begin{table}[t]
    \centering
    \caption{Statistics of the datasets.}
    \vspace{-1em}
    \label{table:dataset}
    \resizebox{1.0\linewidth}{!}{
    \begin{tabular}{l|l|l|l|l|l}
        \toprule
         & \ucr & \iops & \smd & \smap & \msl \\
        \midrule
        \# of Subdatasets & 250 & 29 & 28 & 50 & 23 \\
        \# of Features & 1 & 1 & 38 & 25 & 55 \\
        Domain & Various & Cloud KPIs & Server & \multicolumn{2}{c}{Spacecrafts} \\
        \multicolumn{6}{c}{\textit{Training set (Average per subdataset)}} \\
        Sequence Length & 30522 & 103588 & 25300 & 2752 & 2410 \\
        \# of Anomalies & 0.00 & 42.35 & - & - & - \\
        Anomaly Ratio (\%)& 0.00 & 2.49 & - & - & - \\
        \multicolumn{6}{c}{\textit{Test set (Average per subdataset)}} \\
        Sequence Length & 56173 & 100649 & 25300 & 8203 & 2923 \\
        \# of Anomalies & 1.00 & 50.69 & 11.68 & 1.28 & 1.26 \\
        Anomaly Ratio (\%)& 0.85 & 1.81 & 4.21 & 11.14 & 13.01 \\
        \bottomrule
    \end{tabular}
    }
\vspace{-1em}
\end{table}

\begin{table*}[t]
    \caption{
    Anomaly detection performance.
    V-R and V-P are VUS-ROC and VUS-PR~\cite{vus}.
    RF is Range F-score~\cite{rf}.
    Best results are in \textbf{bold}, and second best results are \underline{underlined} (higher is better).
    }
    \vspace{-1em}
    \label{table:mainresult}
    \centering
    \resizebox{0.91\linewidth}{!}{
  \begin{tabular}{l|llll|lll|lll|lll|lll|lll}
  \toprule
   & \multicolumn{4}{c|}{\ucr} & \multicolumn{3}{c|}{\iops} & \multicolumn{3}{c|}{\smd} & \multicolumn{3}{c|}{\smap} & \multicolumn{3}{c|}{\msl} & \multicolumn{3}{c}{Average} \\
   & V-R & V-P & RF & Acc & V-R & V-P & RF & V-R & V-P & RF & V-R & V-P & RF & V-R & V-P & RF & V-R & V-P & RF \\
  \midrule
\isf&0.641&0.106&0.041&0.236&0.798&0.169&0.061&0.735&0.245&0.093&0.506&0.137&0.059&0.588&0.208&0.080&0.653&0.173&0.067\\
\ocsvm&0.645&0.167&0.092&0.300&0.848&0.336&0.152&0.698&0.239&0.156&$\underline{0.633}$&0.345&0.027&0.697&0.428&0.038&0.704&0.303&0.093\\
\usad&0.613&0.048&0.034&0.272&0.772&0.160&0.026&0.547&0.166&0.077&0.434&0.117&0.092&0.653&0.331&0.073&0.604&0.164&0.060\\
\tranad&0.565&0.020&0.014&0.121&0.647&0.097&0.025&0.312&0.074&0.080&0.416&0.225&0.109&0.408&0.196&0.080&0.469&0.122&0.061\\
\lstmvae&0.739&0.141&0.113&0.217&0.864&0.359&0.185&0.711&0.223&0.190&0.610&0.286&0.127&0.665&0.369&0.176&0.718&0.276&0.158\\
\beatgan&0.746&0.205&0.137&0.351&$\underline{0.889}$&$\mathbf{0.487}$&$\mathbf{0.239}$&0.819&0.356&0.232&0.629&0.309&0.143&0.694&0.395&0.156&$\underline{0.756}$&0.350&0.181\\
\at&0.704&0.132&0.103&0.202&0.881&0.347&0.157&$\mathbf{0.861}$&$\mathbf{0.420}$&$\mathbf{0.252}$&0.616&0.291&0.179&0.670&0.290&0.091&0.746&0.296&0.157\\
\imdiffusion&0.630&0.047&0.028&0.060&0.866&0.358&0.126&$\underline{0.850}$&0.341&0.196&0.610&0.275&0.145&0.678&0.311&0.122&0.727&0.266&0.124\\
\dddr&0.580&0.071&0.060&0.157&0.839&0.396&$\underline{0.213}$&0.828&0.330&$\underline{0.235}$&0.606&0.304&$\underline{0.193}$&$\underline{0.738}$&$\mathbf{0.484}$&$\underline{0.233}$&0.718&0.317&$\underline{0.187}$\\
\ncad&0.548&0.062&0.015&0.232&0.692&0.140&0.016&0.649&0.162&0.044&0.524&0.168&0.021&0.537&0.208&0.000&0.590&0.148&0.019\\
\couta&0.570&0.062&0.056&0.200&0.722&0.187&0.024&0.769&0.268&0.130&0.582&0.258&0.131&0.732&0.376&0.178&0.675&0.230&0.104\\
\capa&$\underline{0.840}$&$\underline{0.441}$&$\mathbf{0.243}$&$\underline{0.613}$&0.834&0.260&0.157&0.697&0.235&0.115&0.621&$\mathbf{0.406}$&0.151&0.629&0.432&0.072&0.724&$\underline{0.355}$&0.148\\
\textbf{\method}&$\mathbf{0.897}$&$\mathbf{0.492}$&$\underline{0.234}$&$\mathbf{0.620}$&$\mathbf{0.911}$&$\underline{0.448}$&0.179&0.813&$\underline{0.366}$&$\underline{0.235}$&$\mathbf{0.642}$&$\underline{0.360}$&$\mathbf{0.196}$&$\mathbf{0.753}$&$\underline{0.482}$&$\mathbf{0.254}$&$\mathbf{0.803}$&$\mathbf{0.430}$&$\mathbf{0.219}$\\
\bottomrule
\end{tabular}
}
\vspace{-1em}
\end{table*}
\subsubsection{Datasets}
We use five datasets in our paper:
(1) UCR Anomaly Archive (\ucr)~\cite{current},
(2) \iops
\footnote{\url{https://competition.aiops-challenge.com/home/competition/1484452272200032281}},
(3) Server Machine Dataset (\smd)~\cite{omnianomaly},
(4) Soil Moisture Active Passive (\smap) satellite~\cite{smapmsl},
and (5) Mars Science Laboratory (\msl)~\cite{smapmsl} rover.
\tabl{\ref{table:dataset}} shows the statistics of each dataset.
\ucr does not contain any anomalies in the training set.
The labels of the training sets in \smd, \smap, and \msl are not provided.
More details about the datasets can be found in \apdx{\ref{apdx:datasets}}. 
We preprocess the datasets by min-max normalization.
For \smap and \msl, we aggregate all the sub-datasets, and for the others, we trained and tested separately for each of the subdatasets, and present the average results of five runs.

\subsubsection{Baselines} 
We compare our method with the following 12 baselines in five categories, including two classical methods, seven normality assumption-based methods, and three binary anomaly assumption-based methods as follows:
(1) Classical machine learning methods (\isf~\cite{isolationforest} and \ocsvm~\cite{ocsvm}),
(2) GAN, VAE, and AE-based methods (\beatgan~\cite{beatgan}, \lstmvae~\cite{lstmvae}, and \usad~\cite{usad}),
(3) Transformer-based methods (\at~\cite{anomalytransformer} and \tranad\cite{tranad}),
(4) Diffusion-based methods (\imdiffusion~\cite{imdiffusion} and \dddr~\cite{dddr}),
and (5) Binary anomaly assumption-based methods (\ncad~\cite{ncad}, \couta~\cite{couta}, and \capa~\cite{cutaddpaste}).
We provide detailed descriptions of all the baselines in \apdx{\ref{apdx:baselines}}.

\subsubsection{Evaluation metrics} 
It is argued that the traditional metric for TSAD, namely the F1 score computed using the point adjustment (PA) protocol, overestimates model performance~\cite{towards,quovadisTAD}.
To ensure a more comprehensive evaluation, we adopt various threshold-independent metrics.
In particular, we employ five range-based measures designed to provide a robust and reliable assessment for TSAD:
Range F-score~\cite{rf}, Range-AUC-ROC, Range-AUC-PR, volume under the surface (VUS)-ROC, and VUS-PR~\cite{vus}.
Of these, VUS-PR is identified as the most reliable and accurate metric~\cite{elephant}.
For Range-AUC-ROC and Range-AUC-PR, we define the buffer region as half the window size.
Additionally, for UCR, we use accuracy as an evaluation metric, where an anomaly is considered detected if the highest anomaly score falls within the given anomaly range.

\subsection{Q1: Effectiveness} \label{sec:q1}
We evaluate the effectiveness of \method.
\tabl{\ref{table:mainresult}} shows the performance of \method and the 12 baseline methods for five datasets.
We include the results of other metrics in \apdx{\ref{apdx:results}}.
Our proposed method \method achieves the highest average detection accuracy across the datasets,
demonstrating improvements of $5.9\%$ in VOC-ROC, $17.4\%$ in VOC-PR, and $14.6\%$ in Range F-score compared with the second-best methods, \beatgan, \capa, and \dddr, respectively.
These results demonstrate that \method, which employs both reconstruction and classification and addresses the limitations of binary anomaly assumption-based methods, is effective for TSAD.
Reconstruction-based methods, specifically \beatgan, \at, \imdiffusion, and \dddr, achieve strong detection performance with simpler datasets, such as \iops and \smd.
However, they face challenges on \ucr, which contains many pattern-wise anomalies, reflecting the limitation of the reconstruction-based method, namely that it is vulnerable to subtle anomalies.
\ncad and \couta, which rely on the binary anomaly assumption, also struggle with \ucr due to their lack of diversity in data augmentation, particularly their inability to generate pattern-wise anomalies.
In contrast, \capa obtains better performance with \ucr thanks to its time series-specific data augmentation, which is capable of mimicking both point-wise and pattern-wise anomalies.
However, \capa underperforms \method across many metrics and datasets because it fails to address contaminated/false anomalies.

\subsection{Q2: Ablation Study} \label{sec:q2}
\begin{figure}[t]
    \centering
    \begin{tabular}{cc}
        \hspace{-.5em}
        \includegraphics[width=0.48\linewidth]{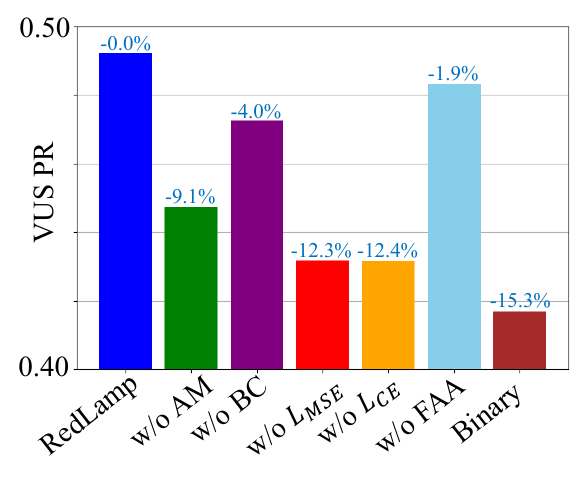} & 
        \includegraphics[width=0.48\linewidth]{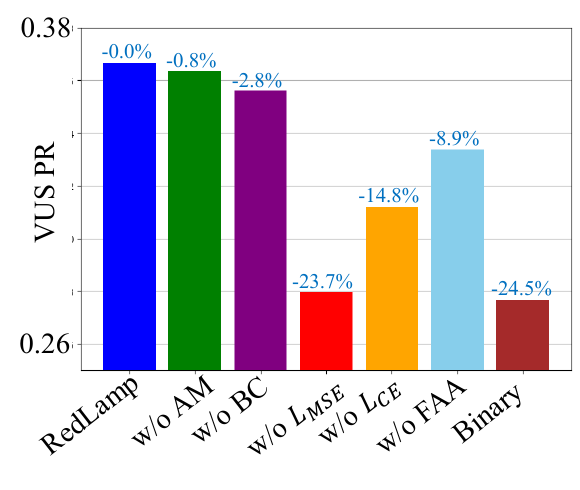} \\[-1em] 
        (a) \ucr & (b) \smd \\ 
    \end{tabular}
    \vspace{-1.em}
    \caption{
        Ablation study results.
    }
    \label{fig:ablation}
    \vspace{-1.em}
\end{figure}
We conduct an ablation study to assess the contribution of each component of our method.
Six ablated variants are used:
(1) w/o anomaly mask (AM): $\masktk{t}{k}=\mathbf{0}$ is applied. 
(2) w/o backward correction (BC): $\probnorm=\probanom=0$ is set.
(3) w/o masked reconstruction loss ($\lossfuncmse$): $\lossweight=1.0$ is used.
(4) w/o cross-entropy loss ($\lossfuncce$): $\lossweight=0.0$ is set.
(5) w/o FAA: $\threshold=1.0$ is applied.
(6) Binary: conducts binary classification, where $11$ data augmentations are used to generate pseudo-anomalies and treat them as the same class.
Considering the imbalanced class problem, the total number of pseudo-anomalies is set at a 1:1 ratio relative to the training set size.
\fig{\ref{fig:ablation}} shows results for \method and its variants.
\method outperforms the six variants, indicating that each component contributes to the detection performance.
AM and BC are particularly effective on UCR, and FAA excels on SMD.
The results of w/o $\lossfuncmse$ and $\lossfuncce$ show the effectiveness of employing both reconstruction and classification.
The comparison between \method and Binary demonstrates that the multiclass anomaly assumption is more effective than the binary anomaly assumption.

\subsection{Q3. Sensitivity} \label{sec:q3}
\begin{figure}[t]
    \centering
    \begin{tabular}{cc} 
        \hspace{-1.5em}
        \includegraphics[height=9em]{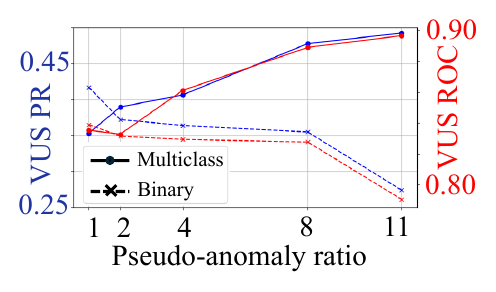} &
        \hspace{-1.5em}
        \includegraphics[height=9em]{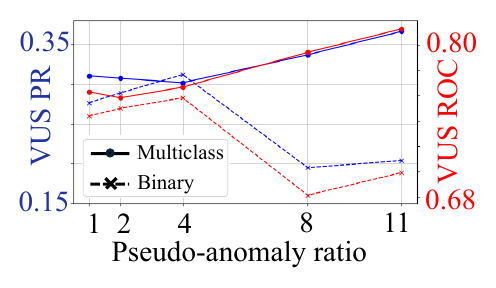} \\ 
        [-0.8em]
        (a) UCR & (b) SMD \\ 
    \end{tabular}
    \vspace{-1.em}
    \caption{
        Comparison of multiclass and binary classification.
    }
    \label{fig:quantity}
    \vspace{-1.em}
\end{figure}

Here, we conduct a sensitivity analysis to study the effect of the key hyperparameter, namely the number of data augmentations.
The investigation of effective data augmentations and other hyperparameter studies are available in \apdx{\ref{apdx:augmentation} and \ref{apdx:hyperparameter}}, respectively.

\myparaitemize{Effect of multiclass pseudo-anomalies}
We investigate the impact of increasing the number of anomaly data augmentations in both multiclass classification and binary classification settings.
For multiclass classification, we vary the number of data augmentations from $1$ to $11$.
We randomly choose the types of pseudo-anomalies in each experiment.
Note that Normal is always included.
Binary classification uses all $11$ types of pseudo-anomalies, and we increase the ratio of the anomaly quantity w.r.t. the training set size.
The results in \fig{\ref{fig:quantity}} indicate that the optimal anomaly quantity ratio for Binary is $1$ for UCR and $4$ for SMD.
At this optimal ratio, Binary outperforms Multiclass, benefiting from the diversity of anomalies.
However, as the ratio increases, Binary experiences performance degradation due to the class imbalance problem.
In contrast, increasing the number of classes in Multiclass improves performance, which can be attributed to the enhanced diversity of anomalies represented in the augmented training set, stressing the importance of filling the diversity gap.

\subsection{Q4. Robustness} \label{sec:q4}
\begin{figure}[t]
    \centering
    \includegraphics[height=12em]{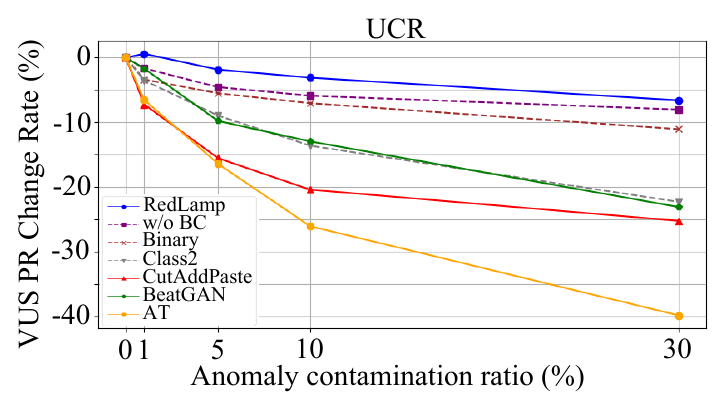} \\
    \vspace{-1.em}
    \caption{
        Robustness w.r.t. anomaly contamination.
    }
    \label{fig:contami}
    \vspace{-1.em}
\end{figure}
\begin{figure*}[t]
    \centering
    \begin{tabular}{c}
        \hspace{-1em}
        \includegraphics[height=29em]{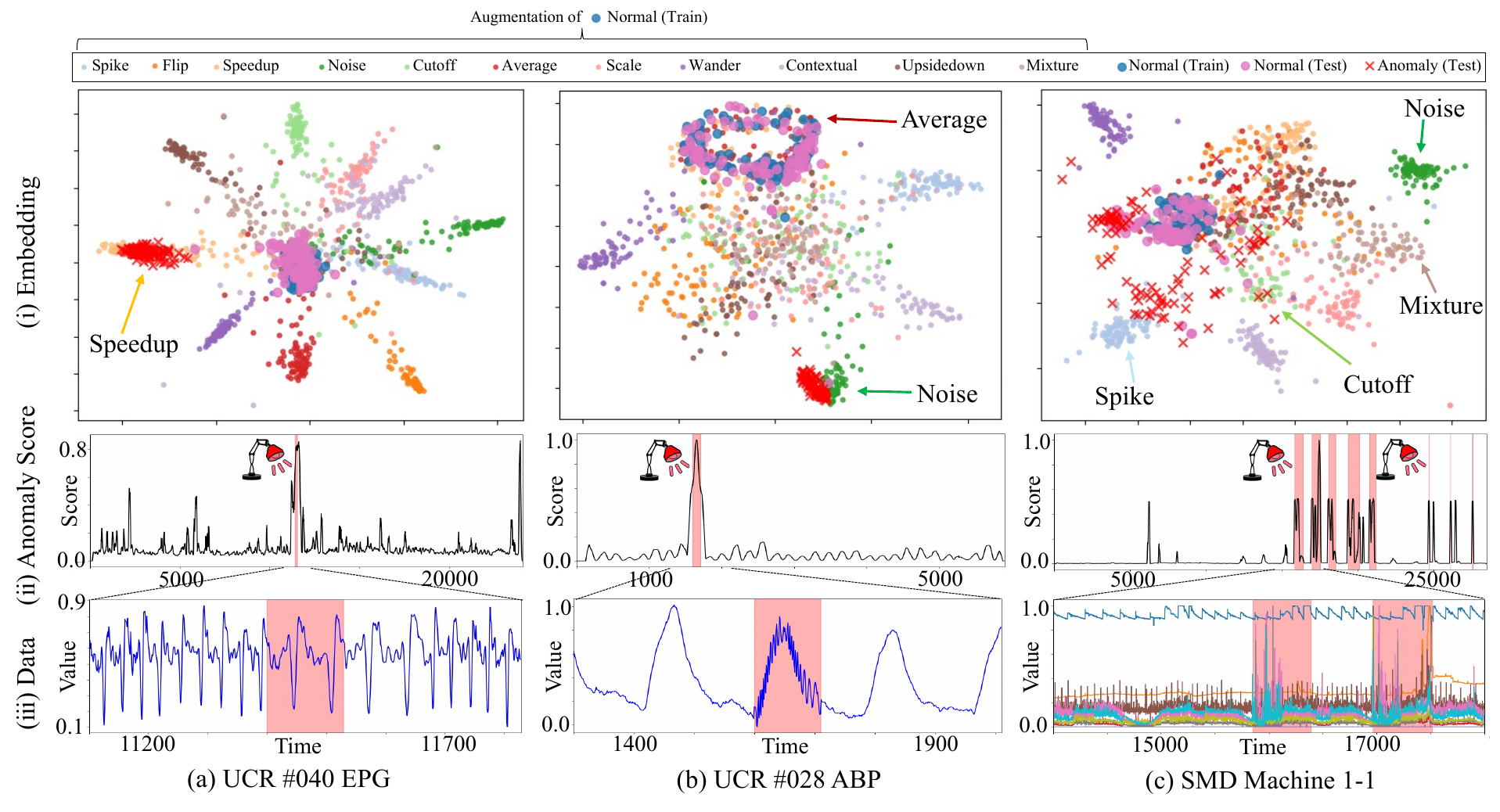} \\
    \end{tabular}
    \vspace{-2.em}
    \caption{
    Visualization of results from (a)(b) \ucr, and (c) \smd datasets.
    (\romanone) Two-dimensional t-SNE plots of the embeddings of the training, pseudo-anomaly, and test instances.
    (\romantwo) Anomaly score of the entire sequence, where anomalous parts are indicated with a red background.
    (\romanthree) Actual data focusing on anomalous parts. For \smd, we show ten features.
    }
    \label{fig:tsne}
    \vspace{-1.em}
\end{figure*}
We evaluate the robustness of \method against anomaly contamination.
We use \ucr for this experiment because the training set does not contain anomalies.
To simulate anomaly contamination, we create a contaminated dataset by randomly mixing $11$ types of anomaly instances into the training set.
We vary the anomaly contamination ratio from $1\%$ to $30\%$
\footnote{
$\text{Anomaly contamination ratio (\%)} = \frac{\text{\# of Anomaly instances}}{\text{\# of Normal instances}} * 100$
}.
\fig{\ref{fig:contami}} shows the performance degradation rate relative to a contamination-free setting ($0\%$) for \method, its ablation variants (dotted line), and three baseline methods.
In addition to the variants used in \secton{\ref{sec:q2}}, (2) w/o BC and (6) Binary, we introduce (7) Class2, which employs a randomly selected single type of pseudo-anomaly while keeping all other aspects identical to Binary.

Normality assumption-based methods, \beatgan and \at, suffer significant performance degradation under contamination. 
This is because they consider all the training samples normal and learn to reconstruct even contaminated anomalies.
As a result, their models become powerful enough to reconstruct even the anomalous parts.
Despite employing data augmentation, \capa also experiences a substantial performance drop.
Specifically, under a $5\%$ anomaly contamination rate, \method exhibits only a $1.8\%$ decrease in performance, whereas \capa suffers a $15.5\%$ decline.
These results underscore the inherent limitations of binary anomaly assumption-based methods --- namely, the diversity gap, false anomalies, and overconfidence in the labels --- which render them vulnerable to anomaly contamination.

The comparison between \method and w/o BC emphasizes the importance of utilizing soft labels to maintain robustness against anomaly contamination, as BC accounts for the probability of anomaly samples being misclassified as normal.
To assess the impact of incorporating diverse data augmentations, we compare Class2 with Binary.
The only difference between these settings lies in the diversity of pseudo-anomalies: Class2 employs only two pseudo-anomalies, whereas Binary utilizes all $11$.
Nevertheless, Class2 exhibits a significant performance drop compared with Binary, which underscores the necessity of bridging the diversity gap to ensure robust performance under anomaly contamination.
\method outperforms Binary across all contamination ratios, which indicates the superiority of the multiclass anomaly assumption over the binary anomaly assumption, even for anomaly contamination.

\subsection{Q5. Explainability} \label{sec:q5}
We demonstrate the explainability of \method.
\fig{\ref{fig:tsne}} is a visualization of (\romanone) embeddings via t-SNE, (\romantwo) anomaly score, and (\romanthree) actual data focusing on anomalous parts.
We provide three examples from (a),(b) \ucr, and (c) \smd.

\myparaitemize{\ucr \#040 EPG}
In \fig{\ref{fig:tsne}} (a) (\romanone), different classes of pseudo-anomalies are clearly separated in the latent space.
The embeddings of \textit{Anomaly (Test)} are gathered around \textit{Speedup}, while \textit{Normal (Test)} overlap \textit{Normal (Train)}.
This matches the actual anomalous parts shown in (\romanthree) that have a longer periodicity than normal, suggesting that the embeddings explain the anomalous parts.

\myparaitemize{\ucr \#028 ABP}
As shown in \fig{\ref{fig:tsne}} (b) (\romanthree), the anomalous parts appear to be the area where the noise is added to the normal peak. 
As we expected, the embeddings of the \textit{Anomaly (Test)} cluster around \textit{Noise}, as shown in (\romanone).
Due to the longer periodicity of the data relative to the local window size ($\timestep=100$), \textit{Average} becomes a pseudo-normal class, resulting in its embeddings being positioned close to \textit{Normal (Train/Test)}.
However, thanks to our adjusted anomaly-class score, \method detects anomalous parts.
As illustrated in (\romantwo), the anomaly score highlights only the anomalous parts.

\myparaitemize{\smd Machine 1-1}
We analyze multivariate dataset \smd in \fig{\ref{fig:tsne}}~(c).
\smd contains multiple anomalies in a sequence, and given its high dimensionality, it is difficult to analyze the anomalies from the visualization of sequences in (\romanthree).
However, \method offers insights into how closely anomalies relate to specific augmentations through the embeddings shown in (\romanone).
The embeddings of \textit{Anomaly (Test)} are widely distributed in the latent space, yet some are located near \textit{Spike} or \textit{Cutoff}.
Additionally, it also reveals that the anomalies are distinct from \textit{Noise} or \textit{Mixture}.
These findings help us understand the nature of anomalies.

Notably, it provides the explainability that enables the categorization of anomaly types that cannot be achieved by binary anomaly assumption-based methods.
Overall, \method offers an explainable latent space and serves as a tool to analyze real anomalies.

\section{Conclusion}
    \label{060conclusions}
    We proposed the first multiclass anomaly assumption-based method, \method, for unsupervised TSAD.
Our method overcomes the issues associated with binary anomaly assumption-based approaches: the diversity gap, false anomalies, and overconfidence in labels.
The core idea of \method is to learn a multiclass boundary using multiclass pseudo-anomalies generated by diverse data augmentations.
The use of multiclass pseudo-anomalies prevents the diversity gap from occurring.
To mitigate the negative impact of anomaly contamination and false anomalies, we conduct multiclass classification via soft labels, preventing the model from being overconfident but ensuring its robustness against them.
Extensive experiments on real-world datasets demonstrated the effectiveness of \method and its robustness against anomaly contamination.
We also showed that the learned latent space is explainable and provides insight into the real anomalies.
Future work will include the analysis of multivariate time series anomalies and the involvement of data augmentations suitable for multivariate time series.
We believe our work lays the foundation for a new approach to TSAD that exploits data augmentation techniques.

\begin{acks}
The authors would like to thank the anonymous referees for their valuable comments and helpful suggestions.
This work was supported by
JST CREST JPMJCR23M3, 
JST COI-NEXT JPMJPF2009, 
JPMJPF2115
.
\end{acks}

\bibliographystyle{ACM-Reference-Format}
\balance
\bibliography{ref_denoising}


\begin{thebibliography}{52}


\ifx \showCODEN    \undefined \def \showCODEN     #1{\unskip}     \fi
\ifx \showDOI      \undefined \def \showDOI       #1{#1}\fi
\ifx \showISBNx    \undefined \def \showISBNx     #1{\unskip}     \fi
\ifx \showISBNxiii \undefined \def \showISBNxiii  #1{\unskip}     \fi
\ifx \showISSN     \undefined \def \showISSN      #1{\unskip}     \fi
\ifx \showLCCN     \undefined \def \showLCCN      #1{\unskip}     \fi
\ifx \shownote     \undefined \def \shownote      #1{#1}          \fi
\ifx \showarticletitle \undefined \def \showarticletitle #1{#1}   \fi
\ifx \showURL      \undefined \def \showURL       {\relax}        \fi
\providecommand\bibfield[2]{#2}
\providecommand\bibinfo[2]{#2}
\providecommand\natexlab[1]{#1}
\providecommand\showeprint[2][]{arXiv:#2}

\bibitem[Abdulaal et~al\mbox{.}(2021)]%
        {asynchronous}
\bibfield{author}{\bibinfo{person}{Ahmed Abdulaal}, \bibinfo{person}{Zhuanghua Liu}, {and} \bibinfo{person}{Tomer Lancewicki}.} \bibinfo{year}{2021}\natexlab{}.
\newblock \showarticletitle{Practical approach to asynchronous multivariate time series anomaly detection and localization}. In \bibinfo{booktitle}{\emph{KDD}}. \bibinfo{pages}{2485--2494}.
\newblock


\bibitem[Audibert et~al\mbox{.}(2020)]%
        {usad}
\bibfield{author}{\bibinfo{person}{Julien Audibert}, \bibinfo{person}{Pietro Michiardi}, \bibinfo{person}{Fr{\'e}d{\'e}ric Guyard}, \bibinfo{person}{S{\'e}bastien Marti}, {and} \bibinfo{person}{Maria~A Zuluaga}.} \bibinfo{year}{2020}\natexlab{}.
\newblock \showarticletitle{Usad: Unsupervised anomaly detection on multivariate time series}. In \bibinfo{booktitle}{\emph{KDD}}. \bibinfo{pages}{3395--3404}.
\newblock


\bibitem[Bashar and Nayak(2020)]%
        {tanogan}
\bibfield{author}{\bibinfo{person}{Md~Abul Bashar} {and} \bibinfo{person}{Richi Nayak}.} \bibinfo{year}{2020}\natexlab{}.
\newblock \showarticletitle{TAnoGAN: Time series anomaly detection with generative adversarial networks}. In \bibinfo{booktitle}{\emph{2020 IEEE Symposium Series on Computational Intelligence (SSCI)}}. IEEE, \bibinfo{pages}{1778--1785}.
\newblock


\bibitem[Bl{\'a}zquez-Garc{\'\i}a et~al\mbox{.}(2021)]%
        {survey_acm}
\bibfield{author}{\bibinfo{person}{Ane Bl{\'a}zquez-Garc{\'\i}a}, \bibinfo{person}{Angel Conde}, \bibinfo{person}{Usue Mori}, {and} \bibinfo{person}{Jose~A Lozano}.} \bibinfo{year}{2021}\natexlab{}.
\newblock \showarticletitle{A review on outlier/anomaly detection in time series data}.
\newblock \bibinfo{journal}{\emph{ACM computing surveys (CSUR)}} \bibinfo{volume}{54}, \bibinfo{number}{3} (\bibinfo{year}{2021}), \bibinfo{pages}{1--33}.
\newblock


\bibitem[Braei and Wagner(2020)]%
        {survey_units}
\bibfield{author}{\bibinfo{person}{Mohammad Braei} {and} \bibinfo{person}{Sebastian Wagner}.} \bibinfo{year}{2020}\natexlab{}.
\newblock \showarticletitle{Anomaly detection in univariate time-series: A survey on the state-of-the-art}.
\newblock \bibinfo{journal}{\emph{arXiv preprint arXiv:2004.00433}} (\bibinfo{year}{2020}).
\newblock


\bibitem[Carmona et~al\mbox{.}(2022)]%
        {ncad}
\bibfield{author}{\bibinfo{person}{Chris~U. Carmona}, \bibinfo{person}{François-Xavier Aubet}, \bibinfo{person}{Valentin Flunkert}, {and} \bibinfo{person}{Jan Gasthaus}.} \bibinfo{year}{2022}\natexlab{}.
\newblock \showarticletitle{Neural Contextual Anomaly Detection for Time Series}. In \bibinfo{booktitle}{\emph{IJCAI}}. \bibinfo{pages}{2843--2851}.
\newblock


\bibitem[Castellani et~al\mbox{.}(2021)]%
        {srea}
\bibfield{author}{\bibinfo{person}{Andrea Castellani}, \bibinfo{person}{Sebastian Schmitt}, {and} \bibinfo{person}{Barbara Hammer}.} \bibinfo{year}{2021}\natexlab{}.
\newblock \showarticletitle{Estimating the electrical power output of industrial devices with end-to-end time-series classification in the presence of label noise}. In \bibinfo{booktitle}{\emph{ECML}}. Springer, \bibinfo{pages}{469--484}.
\newblock


\bibitem[Challu et~al\mbox{.}(2022)]%
        {dghl}
\bibfield{author}{\bibinfo{person}{Cristian~I Challu}, \bibinfo{person}{Peihong Jiang}, \bibinfo{person}{Ying~Nian Wu}, {and} \bibinfo{person}{Laurent Callot}.} \bibinfo{year}{2022}\natexlab{}.
\newblock \showarticletitle{Deep generative model with hierarchical latent factors for time series anomaly detection}. In \bibinfo{booktitle}{\emph{AISTATS}}. PMLR, \bibinfo{pages}{1643--1654}.
\newblock


\bibitem[Chen et~al\mbox{.}(2023)]%
        {imdiffusion}
\bibfield{author}{\bibinfo{person}{Yuhang Chen}, \bibinfo{person}{Chaoyun Zhang}, \bibinfo{person}{Minghua Ma}, \bibinfo{person}{Yudong Liu}, \bibinfo{person}{Ruomeng Ding}, \bibinfo{person}{Bowen Li}, \bibinfo{person}{Shilin He}, \bibinfo{person}{Saravan Rajmohan}, \bibinfo{person}{Qingwei Lin}, {and} \bibinfo{person}{Dongmei Zhang}.} \bibinfo{year}{2023}\natexlab{}.
\newblock \showarticletitle{ImDiffusion: Imputed Diffusion Models for Multivariate Time Series Anomaly Detection}.
\newblock \bibinfo{journal}{\emph{Proc. VLDB Endow.}} \bibinfo{volume}{17}, \bibinfo{number}{3} (\bibinfo{year}{2023}), \bibinfo{pages}{359–372}.
\newblock
\showISSN{2150-8097}


\bibitem[Deng and Hooi(2021)]%
        {gdn}
\bibfield{author}{\bibinfo{person}{Ailin Deng} {and} \bibinfo{person}{Bryan Hooi}.} \bibinfo{year}{2021}\natexlab{}.
\newblock \showarticletitle{Graph neural network-based anomaly detection in multivariate time series}. In \bibinfo{booktitle}{\emph{AAAI}}, Vol.~\bibinfo{volume}{35}. \bibinfo{pages}{4027--4035}.
\newblock


\bibitem[DeVries(2017)]%
        {cutout}
\bibfield{author}{\bibinfo{person}{Terrance DeVries}.} \bibinfo{year}{2017}\natexlab{}.
\newblock \showarticletitle{Improved Regularization of Convolutional Neural Networks with Cutout}.
\newblock \bibinfo{journal}{\emph{arXiv preprint arXiv:1708.04552}} (\bibinfo{year}{2017}).
\newblock


\bibitem[Goswami et~al\mbox{.}(2023a)]%
        {unsupervisedTSAD}
\bibfield{author}{\bibinfo{person}{Mononito Goswami}, \bibinfo{person}{Cristian~Ignacio Challu}, \bibinfo{person}{Laurent Callot}, \bibinfo{person}{Lenon Minorics}, {and} \bibinfo{person}{Andrey Kan}.} \bibinfo{year}{2023}\natexlab{a}.
\newblock \showarticletitle{Unsupervised Model Selection for Time Series Anomaly Detection}. In \bibinfo{booktitle}{\emph{ICLR}}.
\newblock


\bibitem[Goswami et~al\mbox{.}(2023b)]%
        {modelselection}
\bibfield{author}{\bibinfo{person}{Mononito Goswami}, \bibinfo{person}{Cristian~Ignacio Challu}, \bibinfo{person}{Laurent Callot}, \bibinfo{person}{Lenon Minorics}, {and} \bibinfo{person}{Andrey Kan}.} \bibinfo{year}{2023}\natexlab{b}.
\newblock \showarticletitle{Unsupervised Model Selection for Time Series Anomaly Detection}. In \bibinfo{booktitle}{\emph{ICLR}}.
\newblock


\bibitem[Han and Woo(2022)]%
        {fusagnet}
\bibfield{author}{\bibinfo{person}{Siho Han} {and} \bibinfo{person}{Simon~S Woo}.} \bibinfo{year}{2022}\natexlab{}.
\newblock \showarticletitle{Learning sparse latent graph representations for anomaly detection in multivariate time series}. In \bibinfo{booktitle}{\emph{KDD}}. \bibinfo{pages}{2977--2986}.
\newblock


\bibitem[Hendrycks et~al\mbox{.}(2019)]%
        {oe}
\bibfield{author}{\bibinfo{person}{Dan Hendrycks}, \bibinfo{person}{Mantas Mazeika}, {and} \bibinfo{person}{Thomas Dietterich}.} \bibinfo{year}{2019}\natexlab{}.
\newblock \showarticletitle{Deep Anomaly Detection with Outlier Exposure}.
\newblock \bibinfo{journal}{\emph{ICLR}} (\bibinfo{year}{2019}).
\newblock


\bibitem[Hundman et~al\mbox{.}(2018)]%
        {smapmsl}
\bibfield{author}{\bibinfo{person}{Kyle Hundman}, \bibinfo{person}{Valentino Constantinou}, \bibinfo{person}{Christopher Laporte}, \bibinfo{person}{Ian Colwell}, {and} \bibinfo{person}{Tom Soderstrom}.} \bibinfo{year}{2018}\natexlab{}.
\newblock \showarticletitle{Detecting spacecraft anomalies using lstms and nonparametric dynamic thresholding}. In \bibinfo{booktitle}{\emph{KDD}}. \bibinfo{pages}{387--395}.
\newblock


\bibitem[Ismail~Fawaz et~al\mbox{.}(2019)]%
        {ismail2019deep}
\bibfield{author}{\bibinfo{person}{Hassan Ismail~Fawaz}, \bibinfo{person}{Germain Forestier}, \bibinfo{person}{Jonathan Weber}, \bibinfo{person}{Lhassane Idoumghar}, {and} \bibinfo{person}{Pierre-Alain Muller}.} \bibinfo{year}{2019}\natexlab{}.
\newblock \showarticletitle{Deep learning for time series classification: a review}.
\newblock \bibinfo{journal}{\emph{Data mining and knowledge discovery}} \bibinfo{volume}{33}, \bibinfo{number}{4} (\bibinfo{year}{2019}), \bibinfo{pages}{917--963}.
\newblock


\bibitem[Iwana and Uchida(2021)]%
        {augempirical}
\bibfield{author}{\bibinfo{person}{Brian~Kenji Iwana} {and} \bibinfo{person}{Seiichi Uchida}.} \bibinfo{year}{2021}\natexlab{}.
\newblock \showarticletitle{An empirical survey of data augmentation for time series classification with neural networks}.
\newblock \bibinfo{journal}{\emph{Plos one}} \bibinfo{volume}{16}, \bibinfo{number}{7} (\bibinfo{year}{2021}).
\newblock


\bibitem[Jeong et~al\mbox{.}(2023)]%
        {anomalybert}
\bibfield{author}{\bibinfo{person}{Yungi Jeong}, \bibinfo{person}{Eunseok Yang}, \bibinfo{person}{Jung~Hyun Ryu}, \bibinfo{person}{Imseong Park}, {and} \bibinfo{person}{Myungjoo Kang}.} \bibinfo{year}{2023}\natexlab{}.
\newblock \showarticletitle{Anomalybert: Self-supervised transformer for time series anomaly detection using data degradation scheme}.
\newblock \bibinfo{journal}{\emph{arXiv preprint arXiv:2305.04468}} (\bibinfo{year}{2023}).
\newblock


\bibitem[Kim et~al\mbox{.}(2022)]%
        {towards}
\bibfield{author}{\bibinfo{person}{Siwon Kim}, \bibinfo{person}{Kukjin Choi}, \bibinfo{person}{Hyun-Soo Choi}, \bibinfo{person}{Byunghan Lee}, {and} \bibinfo{person}{Sungroh Yoon}.} \bibinfo{year}{2022}\natexlab{}.
\newblock \showarticletitle{Towards a rigorous evaluation of time-series anomaly detection}. In \bibinfo{booktitle}{\emph{AAAI}}, Vol.~\bibinfo{volume}{36}. \bibinfo{pages}{7194--7201}.
\newblock


\bibitem[Lai et~al\mbox{.}(2021)]%
        {tsadtaxonomy}
\bibfield{author}{\bibinfo{person}{Kwei-Herng Lai}, \bibinfo{person}{Daochen Zha}, \bibinfo{person}{Junjie Xu}, \bibinfo{person}{Yue Zhao}, \bibinfo{person}{Guanchu Wang}, {and} \bibinfo{person}{Xia Hu}.} \bibinfo{year}{2021}\natexlab{}.
\newblock \showarticletitle{Revisiting Time Series Outlier Detection: Definitions and Benchmarks}. In \bibinfo{booktitle}{\emph{NeurIPS}}, Vol.~\bibinfo{volume}{1}.
\newblock


\bibitem[Li et~al\mbox{.}(2021a)]%
        {cutpaste}
\bibfield{author}{\bibinfo{person}{Chun-Liang Li}, \bibinfo{person}{Kihyuk Sohn}, \bibinfo{person}{Jinsung Yoon}, {and} \bibinfo{person}{Tomas Pfister}.} \bibinfo{year}{2021}\natexlab{a}.
\newblock \showarticletitle{Cutpaste: Self-supervised learning for anomaly detection and localization}. In \bibinfo{booktitle}{\emph{CVPR}}. \bibinfo{pages}{9664--9674}.
\newblock


\bibitem[Li et~al\mbox{.}(2021b)]%
        {interfusion}
\bibfield{author}{\bibinfo{person}{Zhihan Li}, \bibinfo{person}{Youjian Zhao}, \bibinfo{person}{Jiaqi Han}, \bibinfo{person}{Ya Su}, \bibinfo{person}{Rui Jiao}, \bibinfo{person}{Xidao Wen}, {and} \bibinfo{person}{Dan Pei}.} \bibinfo{year}{2021}\natexlab{b}.
\newblock \showarticletitle{Multivariate time series anomaly detection and interpretation using hierarchical inter-metric and temporal embedding}. In \bibinfo{booktitle}{\emph{KDD}}. \bibinfo{pages}{3220--3230}.
\newblock


\bibitem[Liu et~al\mbox{.}(2008)]%
        {isolationforest}
\bibfield{author}{\bibinfo{person}{Fei~Tony Liu}, \bibinfo{person}{Kai~Ming Ting}, {and} \bibinfo{person}{Zhi-Hua Zhou}.} \bibinfo{year}{2008}\natexlab{}.
\newblock \showarticletitle{Isolation Forest}. In \bibinfo{booktitle}{\emph{ICDM}}. \bibinfo{pages}{413--422}.
\newblock


\bibitem[Liu and Paparrizos(2024)]%
        {elephant}
\bibfield{author}{\bibinfo{person}{Qinghua Liu} {and} \bibinfo{person}{John Paparrizos}.} \bibinfo{year}{2024}\natexlab{}.
\newblock \showarticletitle{The Elephant in the Room: Towards A Reliable Time-Series Anomaly Detection Benchmark}. In \bibinfo{booktitle}{\emph{NeurIPS}}.
\newblock


\bibitem[Lu et~al\mbox{.}(2023)]%
        {refurbishment}
\bibfield{author}{\bibinfo{person}{Yangdi Lu}, \bibinfo{person}{Zhiwei Xu}, {and} \bibinfo{person}{Wenbo He}.} \bibinfo{year}{2023}\natexlab{}.
\newblock \showarticletitle{Rethinking Label Refurbishment: Model Robustness under Label Noise}.
\newblock \bibinfo{journal}{\emph{AAAI}} \bibinfo{volume}{37}, \bibinfo{number}{12} (\bibinfo{year}{2023}), \bibinfo{pages}{15000--15008}.
\newblock


\bibitem[Malhotra et~al\mbox{.}(2016)]%
        {engine_lstmencdec}
\bibfield{author}{\bibinfo{person}{Pankaj Malhotra}, \bibinfo{person}{Anusha Ramakrishnan}, \bibinfo{person}{Gaurangi Anand}, \bibinfo{person}{Lovekesh Vig}, \bibinfo{person}{Puneet Agarwal}, {and} \bibinfo{person}{Gautam Shroff}.} \bibinfo{year}{2016}\natexlab{}.
\newblock \showarticletitle{LSTM-based encoder-decoder for multi-sensor anomaly detection}.
\newblock \bibinfo{journal}{\emph{arXiv preprint arXiv:1607.00148}} (\bibinfo{year}{2016}).
\newblock


\bibitem[Paparrizos et~al\mbox{.}(2022)]%
        {vus}
\bibfield{author}{\bibinfo{person}{John Paparrizos}, \bibinfo{person}{Paul Boniol}, \bibinfo{person}{Themis Palpanas}, \bibinfo{person}{Ruey~S. Tsay}, \bibinfo{person}{Aaron Elmore}, {and} \bibinfo{person}{Michael~J. Franklin}.} \bibinfo{year}{2022}\natexlab{}.
\newblock \showarticletitle{Volume under the surface: a new accuracy evaluation measure for time-series anomaly detection}.
\newblock \bibinfo{journal}{\emph{Proc. VLDB Endow.}} \bibinfo{volume}{15}, \bibinfo{number}{11} (\bibinfo{date}{jul} \bibinfo{year}{2022}), \bibinfo{pages}{2774–2787}.
\newblock
\showISSN{2150-8097}


\bibitem[Park et~al\mbox{.}(2018)]%
        {lstmvae}
\bibfield{author}{\bibinfo{person}{Daehyung Park}, \bibinfo{person}{Yuuna Hoshi}, {and} \bibinfo{person}{Charles~C Kemp}.} \bibinfo{year}{2018}\natexlab{}.
\newblock \showarticletitle{A multimodal anomaly detector for robot-assisted feeding using an lstm-based variational autoencoder}.
\newblock \bibinfo{journal}{\emph{IEEE Robotics and Automation Letters}} \bibinfo{volume}{3}, \bibinfo{number}{3} (\bibinfo{year}{2018}), \bibinfo{pages}{1544--1551}.
\newblock


\bibitem[Park et~al\mbox{.}(2017)]%
        {robot}
\bibfield{author}{\bibinfo{person}{Daehyung Park}, \bibinfo{person}{Hokeun Kim}, \bibinfo{person}{Yuuna Hoshi}, \bibinfo{person}{Zackory Erickson}, \bibinfo{person}{Ariel Kapusta}, {and} \bibinfo{person}{Charles~C Kemp}.} \bibinfo{year}{2017}\natexlab{}.
\newblock \showarticletitle{A multimodal execution monitor with anomaly classification for robot-assisted feeding}. In \bibinfo{booktitle}{\emph{2017 IEEE/RSJ International Conference on Intelligent Robots and Systems (IROS)}}. IEEE, \bibinfo{pages}{5406--5413}.
\newblock


\bibitem[Patrini et~al\mbox{.}(2017)]%
        {backward}
\bibfield{author}{\bibinfo{person}{Giorgio Patrini}, \bibinfo{person}{Alessandro Rozza}, \bibinfo{person}{Aditya Krishna~Menon}, \bibinfo{person}{Richard Nock}, {and} \bibinfo{person}{Lizhen Qu}.} \bibinfo{year}{2017}\natexlab{}.
\newblock \showarticletitle{Making deep neural networks robust to label noise: A loss correction approach}. In \bibinfo{booktitle}{\emph{CVPR}}. \bibinfo{pages}{1944--1952}.
\newblock


\bibitem[Sarfraz et~al\mbox{.}(2024)]%
        {quovadisTAD}
\bibfield{author}{\bibinfo{person}{M~Saquib Sarfraz}, \bibinfo{person}{Mei-Yen Chen}, \bibinfo{person}{Lukas Layer}, \bibinfo{person}{Kunyu Peng}, {and} \bibinfo{person}{Marios Koulakis}.} \bibinfo{year}{2024}\natexlab{}.
\newblock \showarticletitle{Position: Quo Vadis, Unsupervised Time Series Anomaly Detection?}. In \bibinfo{booktitle}{\emph{ICML}}.
\newblock


\bibitem[Schmidl et~al\mbox{.}(2022)]%
        {survey_vldb}
\bibfield{author}{\bibinfo{person}{Sebastian Schmidl}, \bibinfo{person}{Phillip Wenig}, {and} \bibinfo{person}{Thorsten Papenbrock}.} \bibinfo{year}{2022}\natexlab{}.
\newblock \showarticletitle{Anomaly Detection in Time Series: A Comprehensive Evaluation}.
\newblock \bibinfo{journal}{\emph{VLDB}} \bibinfo{volume}{15}, \bibinfo{number}{9} (\bibinfo{year}{2022}), \bibinfo{pages}{1779--1797}.
\newblock


\bibitem[Sch{\"o}lkopf et~al\mbox{.}(1999)]%
        {ocsvm}
\bibfield{author}{\bibinfo{person}{Bernhard Sch{\"o}lkopf}, \bibinfo{person}{Robert~C Williamson}, \bibinfo{person}{Alex Smola}, \bibinfo{person}{John Shawe-Taylor}, {and} \bibinfo{person}{John Platt}.} \bibinfo{year}{1999}\natexlab{}.
\newblock \showarticletitle{Support vector method for novelty detection}.
\newblock \bibinfo{journal}{\emph{NeurIPS}}  \bibinfo{volume}{12} (\bibinfo{year}{1999}).
\newblock


\bibitem[Su et~al\mbox{.}(2019)]%
        {omnianomaly}
\bibfield{author}{\bibinfo{person}{Ya Su}, \bibinfo{person}{Youjian Zhao}, \bibinfo{person}{Chenhao Niu}, \bibinfo{person}{Rong Liu}, \bibinfo{person}{Wei Sun}, {and} \bibinfo{person}{Dan Pei}.} \bibinfo{year}{2019}\natexlab{}.
\newblock \showarticletitle{Robust anomaly detection for multivariate time series through stochastic recurrent neural network}. In \bibinfo{booktitle}{\emph{KDD}}. \bibinfo{pages}{2828--2837}.
\newblock


\bibitem[Tan et~al\mbox{.}(2011)]%
        {hs-tree}
\bibfield{author}{\bibinfo{person}{Swee~Chuan Tan}, \bibinfo{person}{Kai~Ming Ting}, {and} \bibinfo{person}{Tony~Fei Liu}.} \bibinfo{year}{2011}\natexlab{}.
\newblock \showarticletitle{Fast anomaly detection for streaming data}. In \bibinfo{booktitle}{\emph{AAAI}} \emph{(\bibinfo{series}{IJCAI'11})}. \bibinfo{publisher}{AAAI Press}, \bibinfo{pages}{1511–1516}.
\newblock
\showISBNx{9781577355144}


\bibitem[Tatbul et~al\mbox{.}(2018)]%
        {rf}
\bibfield{author}{\bibinfo{person}{Nesime Tatbul}, \bibinfo{person}{Tae~Jun Lee}, \bibinfo{person}{Stan Zdonik}, \bibinfo{person}{Mejbah Alam}, {and} \bibinfo{person}{Justin Gottschlich}.} \bibinfo{year}{2018}\natexlab{}.
\newblock \showarticletitle{Precision and recall for time series}. In \bibinfo{booktitle}{\emph{NeurIPS}}. \bibinfo{pages}{1924–1934}.
\newblock


\bibitem[Tuli et~al\mbox{.}(2022)]%
        {tranad}
\bibfield{author}{\bibinfo{person}{Shreshth Tuli}, \bibinfo{person}{Giuliano Casale}, {and} \bibinfo{person}{Nicholas~R Jennings}.} \bibinfo{year}{2022}\natexlab{}.
\newblock \showarticletitle{{TranAD: Deep Transformer Networks for Anomaly Detection in Multivariate Time Series Data}}.
\newblock \bibinfo{journal}{\emph{Proceedings of VLDB}} \bibinfo{volume}{15}, \bibinfo{number}{6} (\bibinfo{year}{2022}), \bibinfo{pages}{1201--1214}.
\newblock


\bibitem[Wang et~al\mbox{.}(2024b)]%
        {dddr}
\bibfield{author}{\bibinfo{person}{Chengsen Wang}, \bibinfo{person}{Zirui Zhuang}, \bibinfo{person}{Qi Qi}, \bibinfo{person}{Jingyu Wang}, \bibinfo{person}{Xingyu Wang}, \bibinfo{person}{Haifeng Sun}, {and} \bibinfo{person}{Jianxin Liao}.} \bibinfo{year}{2024}\natexlab{b}.
\newblock \showarticletitle{Drift doesn't matter: dynamic decomposition with diffusion reconstruction for unstable multivariate time series anomaly detection}.
\newblock \bibinfo{journal}{\emph{NeurIPS}}  \bibinfo{volume}{36} (\bibinfo{year}{2024}).
\newblock


\bibitem[Wang et~al\mbox{.}(2023)]%
        {coca}
\bibfield{author}{\bibinfo{person}{Rui Wang}, \bibinfo{person}{Chongwei Liu}, \bibinfo{person}{Xudong Mou}, \bibinfo{person}{Kai Gao}, \bibinfo{person}{Xiaohui Guo}, \bibinfo{person}{Pin Liu}, \bibinfo{person}{Tianyu Wo}, {and} \bibinfo{person}{Xudong Liu}.} \bibinfo{year}{2023}\natexlab{}.
\newblock \showarticletitle{Deep contrastive one-class time series anomaly detection}. In \bibinfo{booktitle}{\emph{SDM}}. SIAM, \bibinfo{pages}{694--702}.
\newblock


\bibitem[Wang et~al\mbox{.}(2024a)]%
        {cutaddpaste}
\bibfield{author}{\bibinfo{person}{Rui Wang}, \bibinfo{person}{Xudong Mou}, \bibinfo{person}{Renyu Yang}, \bibinfo{person}{Kai Gao}, \bibinfo{person}{Pin Liu}, \bibinfo{person}{Chongwei Liu}, \bibinfo{person}{Tianyu Wo}, {and} \bibinfo{person}{Xudong Liu}.} \bibinfo{year}{2024}\natexlab{a}.
\newblock \showarticletitle{CutAddPaste: Time Series Anomaly Detection by Exploiting Abnormal Knowledge}. In \bibinfo{booktitle}{\emph{KDD}}. \bibinfo{pages}{3176--3187}.
\newblock


\bibitem[Wen et~al\mbox{.}(2021)]%
        {augsurvey}
\bibfield{author}{\bibinfo{person}{Qingsong Wen}, \bibinfo{person}{Liang Sun}, \bibinfo{person}{Fan Yang}, \bibinfo{person}{Xiaomin Song}, \bibinfo{person}{Jingkun Gao}, \bibinfo{person}{Xue Wang}, {and} \bibinfo{person}{Huan Xu}.} \bibinfo{year}{2021}\natexlab{}.
\newblock \showarticletitle{Time Series Data Augmentation for Deep Learning: A Survey}. In \bibinfo{booktitle}{\emph{IJCAI}}. \bibinfo{pages}{4653--4660}.
\newblock
\newblock
\shownote{Survey Track}.


\bibitem[Wu and Keogh(2021)]%
        {current}
\bibfield{author}{\bibinfo{person}{Renjie Wu} {and} \bibinfo{person}{Eamonn~J Keogh}.} \bibinfo{year}{2021}\natexlab{}.
\newblock \showarticletitle{Current time series anomaly detection benchmarks are flawed and are creating the illusion of progress}.
\newblock \bibinfo{journal}{\emph{IEEE transactions on knowledge and data engineering}} \bibinfo{volume}{35}, \bibinfo{number}{3} (\bibinfo{year}{2021}), \bibinfo{pages}{2421--2429}.
\newblock


\bibitem[Wu et~al\mbox{.}(2022)]%
        {mitigatecontamination}
\bibfield{author}{\bibinfo{person}{Shuang Wu}, \bibinfo{person}{Jingyu Zhao}, {and} \bibinfo{person}{Guangjian Tian}.} \bibinfo{year}{2022}\natexlab{}.
\newblock \showarticletitle{Understanding and Mitigating Data Contamination in Deep Anomaly Detection: A Kernel-based Approach}. In \bibinfo{booktitle}{\emph{IJCAI}}. \bibinfo{pages}{2319--2325}.
\newblock


\bibitem[Xiao et~al\mbox{.}(2023)]%
        {diffad}
\bibfield{author}{\bibinfo{person}{Chunjing Xiao}, \bibinfo{person}{Zehua Gou}, \bibinfo{person}{Wenxin Tai}, \bibinfo{person}{Kunpeng Zhang}, {and} \bibinfo{person}{Fan Zhou}.} \bibinfo{year}{2023}\natexlab{}.
\newblock \showarticletitle{Imputation-based time-series anomaly detection with conditional weight-incremental diffusion models}. In \bibinfo{booktitle}{\emph{KDD}}. \bibinfo{pages}{2742--2751}.
\newblock


\bibitem[Xu et~al\mbox{.}(2024)]%
        {couta}
\bibfield{author}{\bibinfo{person}{Hongzuo Xu}, \bibinfo{person}{Yijie Wang}, \bibinfo{person}{Songlei Jian}, \bibinfo{person}{Qing Liao}, \bibinfo{person}{Yongjun Wang}, {and} \bibinfo{person}{Guansong Pang}.} \bibinfo{year}{2024}\natexlab{}.
\newblock \showarticletitle{Calibrated one-class classification for unsupervised time series anomaly detection}.
\newblock \bibinfo{journal}{\emph{IEEE Transactions on Knowledge and Data Engineering}} (\bibinfo{year}{2024}).
\newblock


\bibitem[Xu et~al\mbox{.}(2022)]%
        {anomalytransformer}
\bibfield{author}{\bibinfo{person}{Jiehui Xu}, \bibinfo{person}{Haixu Wu}, \bibinfo{person}{Jianmin Wang}, {and} \bibinfo{person}{Mingsheng Long}.} \bibinfo{year}{2022}\natexlab{}.
\newblock \showarticletitle{Anomaly Transformer: Time Series Anomaly Detection with Association Discrepancy}. In \bibinfo{booktitle}{\emph{ICLR}}.
\newblock


\bibitem[Yang et~al\mbox{.}(2023)]%
        {dcdetector}
\bibfield{author}{\bibinfo{person}{Yiyuan Yang}, \bibinfo{person}{Chaoli Zhang}, \bibinfo{person}{Tian Zhou}, \bibinfo{person}{Qingsong Wen}, {and} \bibinfo{person}{Liang Sun}.} \bibinfo{year}{2023}\natexlab{}.
\newblock \showarticletitle{Dcdetector: Dual attention contrastive representation learning for time series anomaly detection}. In \bibinfo{booktitle}{\emph{KDD}}. \bibinfo{pages}{3033--3045}.
\newblock


\bibitem[Yoo et~al\mbox{.}(2023)]%
        {hyperparameteraug}
\bibfield{author}{\bibinfo{person}{Jaemin Yoo}, \bibinfo{person}{Tiancheng Zhao}, {and} \bibinfo{person}{Leman Akoglu}.} \bibinfo{year}{2023}\natexlab{}.
\newblock \showarticletitle{Data Augmentation is a Hyperparameter: Cherry-picked Self-Supervision for Unsupervised Anomaly Detection is Creating the Illusion of Success}.
\newblock \bibinfo{journal}{\emph{TMLR}} (\bibinfo{year}{2023}).
\newblock
\showISSN{2835-8856}


\bibitem[Yoon et~al\mbox{.}(2022)]%
        {pseudoanomaly}
\bibfield{author}{\bibinfo{person}{Jinsung Yoon}, \bibinfo{person}{Kihyuk Sohn}, \bibinfo{person}{Chun-Liang Li}, \bibinfo{person}{Sercan~O Arik}, \bibinfo{person}{Chen-Yu Lee}, {and} \bibinfo{person}{Tomas Pfister}.} \bibinfo{year}{2022}\natexlab{}.
\newblock \showarticletitle{Self-supervise, Refine, Repeat: Improving Unsupervised Anomaly Detection}.
\newblock \bibinfo{journal}{\emph{TMLR}} (\bibinfo{year}{2022}).
\newblock
\showISSN{2835-8856}


\bibitem[Zavrtanik et~al\mbox{.}(2021)]%
        {dream}
\bibfield{author}{\bibinfo{person}{Vitjan Zavrtanik}, \bibinfo{person}{Matej Kristan}, {and} \bibinfo{person}{Danijel Skocaj}.} \bibinfo{year}{2021}\natexlab{}.
\newblock \showarticletitle{DRAEM - A Discriminatively Trained Reconstruction Embedding for Surface Anomaly Detection}. In \bibinfo{booktitle}{\emph{ICCV}}. \bibinfo{pages}{8330--8339}.
\newblock


\bibitem[Zhou et~al\mbox{.}(2019)]%
        {beatgan}
\bibfield{author}{\bibinfo{person}{Bin Zhou}, \bibinfo{person}{Shenghua Liu}, \bibinfo{person}{Bryan Hooi}, \bibinfo{person}{Xueqi Cheng}, {and} \bibinfo{person}{Jing Ye}.} \bibinfo{year}{2019}\natexlab{}.
\newblock \showarticletitle{Beatgan: Anomalous rhythm detection using adversarially generated time series.}. In \bibinfo{booktitle}{\emph{IJCAI}}, Vol.~\bibinfo{volume}{2019}. \bibinfo{pages}{4433--4439}.
\newblock


\end{thebibliography}

\appendix
\label{100appendix}

\section{Datasets} \label{apdx:datasets}
We provide a detailed description of the datasets used in the experiments.

\myparaitemize{UCR Anomaly Archive (\ucr)~\cite{current}}
This archive comprises 250 diverse univariate time series spanning various domains.
The are categorized into nine groups based on their domain~\cite{modelselection},
(1) Acceleration, (2) Air Temperature, (3) Arterial Blood Pressure (ABP), (4) Electrical Penetration Graph (EPG), (5) Electrocardiogram (ECG), (6) Gait, (7) NASA, (8) Power Demand, and (9) Respiration (RESP).
Each sequence contains a single, carefully validated anomaly, addressing some of the flaws in previously used benchmarks~\cite{current}.

\myparaitemize{\iops\footnote{\url{https://competition.aiops-challenge.com/home/competition/1484452272200032281}}}
This dataset consists of 29 univariate time series representing Key Performance Indicators (KPIs), which are monitored to ensure the stability of web services provided by major Internet companies.

\myparaitemize{Server Machine Dataset (\smd)~\cite{omnianomaly}}
This dataset contains multivariate time series data with 38 features collected from 28 server machines of a large internet company over a five-week period.

\myparaitemize{Soil Moisture Active Passive (\smap) satellite and Mars Science Laboratory (\msl) rover~\cite{smapmsl}}
\smap and \msl are spacecraft datasets collected from NASA. 
\smap measures soil moisture and freeze-thaw states on the Earth's surface using the Mars rover.
\msl collects sensor and actuator data from the Mars rover itself.

For \iops, \smd, \smap, and \msl, the window step was set to 10. For \ucr, due to the high redundancy of repeating patterns, the window step was set within the range of $\{1, 10, 100\}$ to ensure the number of training samples stays below $10000$. Therefore, the number of training samples is given by $\frac{\text{sequence length}}{\text{window step}}$, and some of them overlap.
Although the use of multivariate time series datasets (\smd, \smap, and \msl) generated both support~\cite{dghl,modelselection} and opposition~\cite{current}, we use them because of their multivariate nature.

\section{Baselines} \label{apdx:baselines}
We detail the baselines below.
\begin{itemize}
    \item \isf~\cite{isolationforest}: constructs binary trees based on random space splitting.
    The nodes with shorter path lengths to the root are more likely to be anomalies. 
    \item \ocsvm~\cite{ocsvm}: uses the smallest possible hypersphere to encompass instances.
    The distance from the hypersphere to data located outside the hypersphere is used to determine whether or not the data are anomalous. 
    \item \usad~\cite{usad}: presents an autoencoder architecture whose adversarial-style learning is also inspired by GAN. 
    \item \tranad~\cite{tranad}: is a Transformer-based model that uses attention-based sequence encoders to perform inferences with broader temporal trends, focusing on score-based self-conditioning for robust multi-modal feature extraction and adversarial training. 
    \item \lstmvae~\cite{lstmvae}: is an LSTM-based VAE capturing complex temporal dependencies that models the data-generating process from the latent space to the observed space, which is trained using variational techniques.
    \item \beatgan~\cite{beatgan}: is based on a GAN framework where reconstructions produced by the generator are regularized by the discriminator instead of fixed reconstruction loss functions.
    \item \at~\cite{anomalytransformer}: combines series and prior association to make anomalies distinctive. 
    \item \imdiffusion~\cite{imdiffusion}: integrates the imputation method with a grating masking strategy and utilizes diffusion models to generate masked sequences.
    \item \dddr~\cite{dddr}: tackles the drift via decomposition and reconstruction, overcoming the limitation of the local sliding window.
    \item \ncad~\cite{ncad}: adapts OE, point outlier, and window mixup to generate pseudo-anomalies and then applies the hypersphere classifier discerning boundary between the normal and anomaly classes.
    \item \couta~\cite{couta}: creates pseudo-anomalies and conducts classifications. It adaptively penalizes potential anomalies in the training set that have uncertain predictions to tackle anomaly contamination.
    \item \capa~\cite{cutaddpaste}: focuses on generating diverse time series-specific anomalies. It then applies a TCN-based binary classifier to learn the decision boundary.
\end{itemize}
The baselines are implemented based on previously reported hyperparameters.
The batch size is 16, and the window size is 100.
Similar to \method, we use $10\%$ of the training samples as validation and apply early stopping.
While some methods, such as \at, introduce their own anomaly scoring mechanisms, we use reconstruction error as the anomaly score because it yields better accuracy across the board.
This is because of the inappropriate metrics used in their experiments.
To handle sequential information, we apply the moving average of half of the window size (50 timesteps) to the anomaly score.

\section{Implementation Details} \label{apdx:implementation}
\begin{table}[ht]
    \centering
    \small
    \caption{
    Key hyperparameters for \method.
    }
    \vspace{-1em}
    \label{table:hyperparameter}
    \begin{tabular}{l|p{2cm}}
        \toprule
        Design Parameter & Value \\
        \midrule
        Filter size of convolution layer & [128,128,256,256] \\
        Stride & 2 \\
        Dropout rate & 0.2 \\
        Encoder embedding dimension & 128 \\
        Classifier embedding dimension & 32 \\
        \midrule
        \# of data augmentations: $\nclass$ & 12 \\
        Probability of normal: $\probnorm$ & 0.1 \\
        Probability of anomaly: $\probanom$ & 0.01 \\
        Loss weightage: $\lossweight$ & 0.1 \\
        FAA threshold: $\threshold$ & 0.05 \\
        \bottomrule
    \end{tabular}
\end{table}

\myparaitemize{Architecture}
Our architecture is based on SREA~\cite{srea}.
The encoder is composed of 4 convolutional blocks, subsequently a maxpooling layer and a 1D-convolution layer.
Each block contains a 1D-convolution layer, batch normalization, ReLU activation, and a dropout layer.
The decoder has a symmetric structure to the encoder, with a linear layer for upsampling, 4 convolutional blocks, and a 1D-convolution layer.
The classifier includes a linear layer, batch normalization, ReLU activation, a dropout layer, a linear layer, and a softmax function.

\myparaitemize{Hyperparameter}
We train the proposed model using AdamW with a learning rate of $10^{-3}$.
The batch size is 128, and the training includes early stopping for a maximum of 100 epochs.
The window size is 100.
Other key hyperparameters related to the model architecture are listed in \tabl{\ref{table:hyperparameter}}.
We use consistent hyperparameters throughout all datasets.

\begin{table*}[ht]
    \caption{
    Anomaly detection results.
    R-R and R-P are Range-AUC-ROC and Range-AUC-PR~\cite{vus}.
    }
    \vspace{-1em}
    \label{table:raucresult}
    \centering
    \fontsize{7pt}{7pt}\selectfont
  \begin{tabular}{l|ll|ll|ll|ll|ll|ll}
  \toprule
   & \multicolumn{2}{c|}{\ucr} & \multicolumn{2}{c|}{\iops} & \multicolumn{2}{c|}{\smd} & \multicolumn{2}{c|}{\smap} & \multicolumn{2}{c|}{\msl} & \multicolumn{2}{c}{Average} \\
   & R-R & R-P  & R-R & R-P  & R-R & R-P  & R-R & R-P  & R-R & R-P  & R-R & R-P   \\
  \midrule
\isf&0.653&0.114&0.825&0.217&0.764&0.275&0.532&0.144&0.614&0.223&0.678&0.195\\
\ocsvm&0.662&0.187&0.874&0.463&0.727&0.274&$\underline{0.648}$&0.356&0.712&0.463&0.725&0.348\\
\usad&0.657&0.061&0.840&0.289&0.586&0.195&0.473&0.125&0.673&0.347&0.646&0.204\\
\tranad&0.607&0.025&0.740&0.185&0.380&0.102&0.460&0.245&0.456&0.208&0.529&0.153\\
\lstmvae&0.745&0.140&0.881&0.419&0.732&0.239&0.631&0.280&0.684&0.372&0.735&0.290\\
\beatgan&0.752&0.209&$\underline{0.912}$&$\underline{0.594}$&0.829&0.383&0.643&0.307&0.709&0.398&$\underline{0.769}$&0.378\\
\at&0.720&0.138&0.901&0.442&$\mathbf{0.874}$&$\mathbf{0.457}$&0.629&0.297&0.689&0.311&0.763&0.329\\
\imdiffusion&0.651&0.050&0.889&0.435&$\underline{0.865}$&0.367&0.628&0.276&0.699&0.329&0.746&0.291\\
\dddr&0.610&0.074&0.871&0.492&0.846&0.368&0.622&0.306&0.748&$\underline{0.501}$&0.739&0.348\\
\ncad&0.603&0.084&0.805&0.243&0.722&0.213&0.554&0.178&0.567&0.235&0.650&0.190\\
\couta&0.627&0.087&0.856&0.342&0.834&0.360&0.620&0.283&$\underline{0.753}$&0.414&0.738&0.297\\
\capa&$\underline{0.850}$&$\underline{0.466}$&0.890&0.393&0.738&0.294&0.632&$\mathbf{0.422}$&0.643&0.460&0.751&$\underline{0.407}$\\
\textbf{\method}&$\mathbf{0.902}$&$\mathbf{0.517}$&$\mathbf{0.940}$&$\mathbf{0.636}$&0.837&$\underline{0.429}$&$\mathbf{0.658}$&$\underline{0.374}$&$\mathbf{0.766}$&$\mathbf{0.507}$&$\mathbf{0.821}$&$\mathbf{0.493}$\\
\bottomrule
\end{tabular}
\vspace{-1em}
\end{table*}

\begin{figure}[t]
    \centering
    \hspace{-1.em}
    \begin{tabular}{c}
        \includegraphics[height=8em]{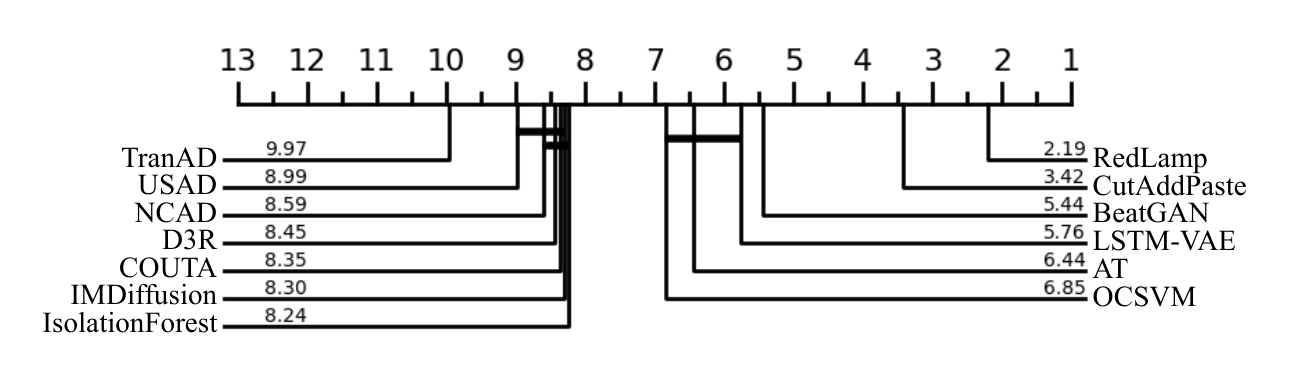} \\[-1em]
        (a) Univariate time series \\
        \includegraphics[height=8em]{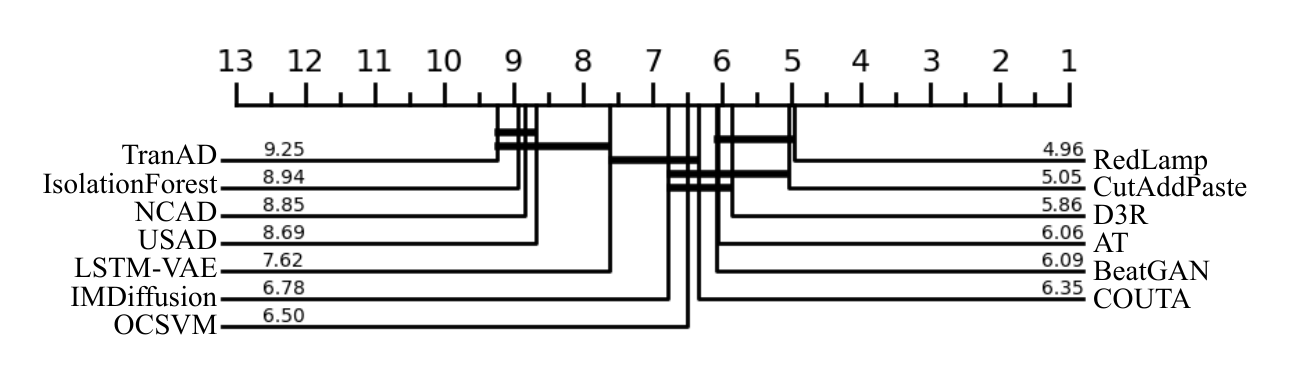} \\[-1em]
        (b) Multivariate time series \\
    \end{tabular}
    \vspace{-1.em}
    \caption{
        Critical difference diagram of VUS-PR.
    }
    \label{fig:cd}
    \vspace{-1.em}
\end{figure}

\myparaitemize{Data augmentation} 
Previous binary anomaly assumption-based methods~\cite{cutaddpaste,coca} require the extensive hyperparameter tuning of data augmentation, such as a scaling factor and the level of trend shifts, for different datasets to prevent false anomalies and the diversity gap from occurring.
While our data augmentations can include hyperparameters, throughout this paper, we generate augmented instances using the consistent data augmentations described in \secton{\ref{sec:augmentation}}, which involve some randomness.

\section{Results of Various Metrics} \label{apdx:results}
We show the corresponding results for other range-based metrics, Range-AUC-ROC and Range-AUC-PR, in \tabl{\ref{table:raucresult}}.
Overall, we observe that \method outperforms the baselines in terms of the average results.

\fig{\ref{fig:cd}} shows the corresponding critical difference diagram of VUS-PR for (a) univariate time series and (b) multivariate time series based on the Wilcoxon-Holm method~\cite{ismail2019deep}, where methods that are not connected by a bold line are significantly different in average rank.
This confirms that in the univariate setting, \method significantly outperforms other methods in average rank.

\section{Effective Data Augmentations} \label{apdx:augmentation}
\begin{figure}[t]
    \centering
    \hspace{-2.em}
    \begin{tabular}{c}
        \includegraphics[height=12em]{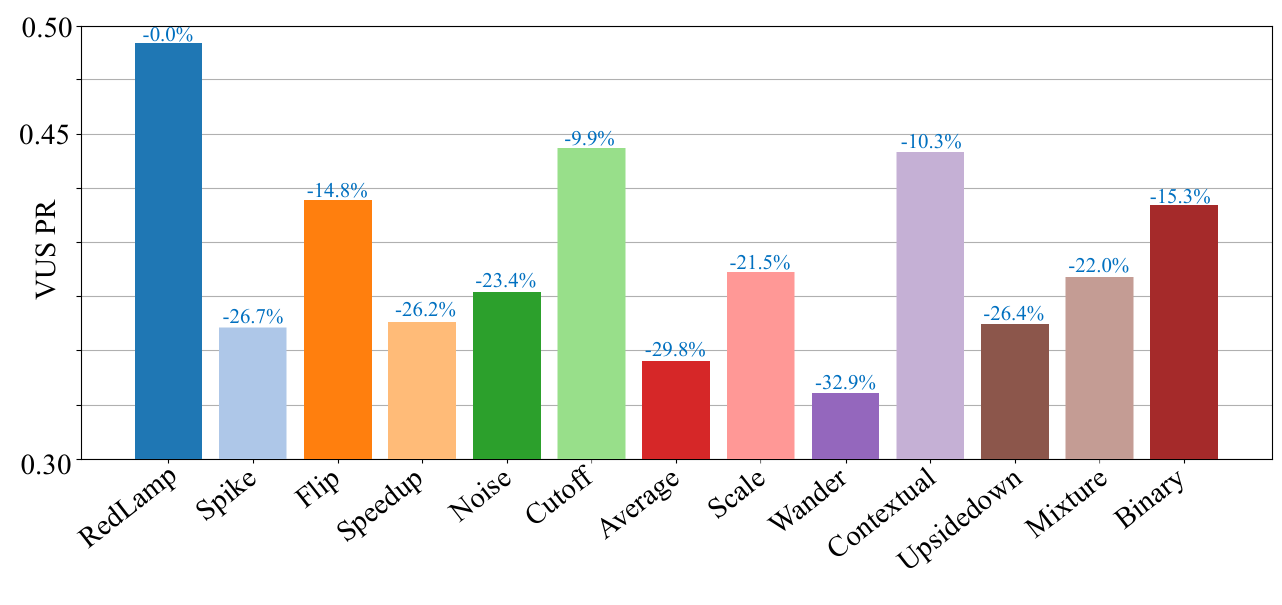} \\[-1em]
        (a) \ucr \\
        \includegraphics[height=12em]{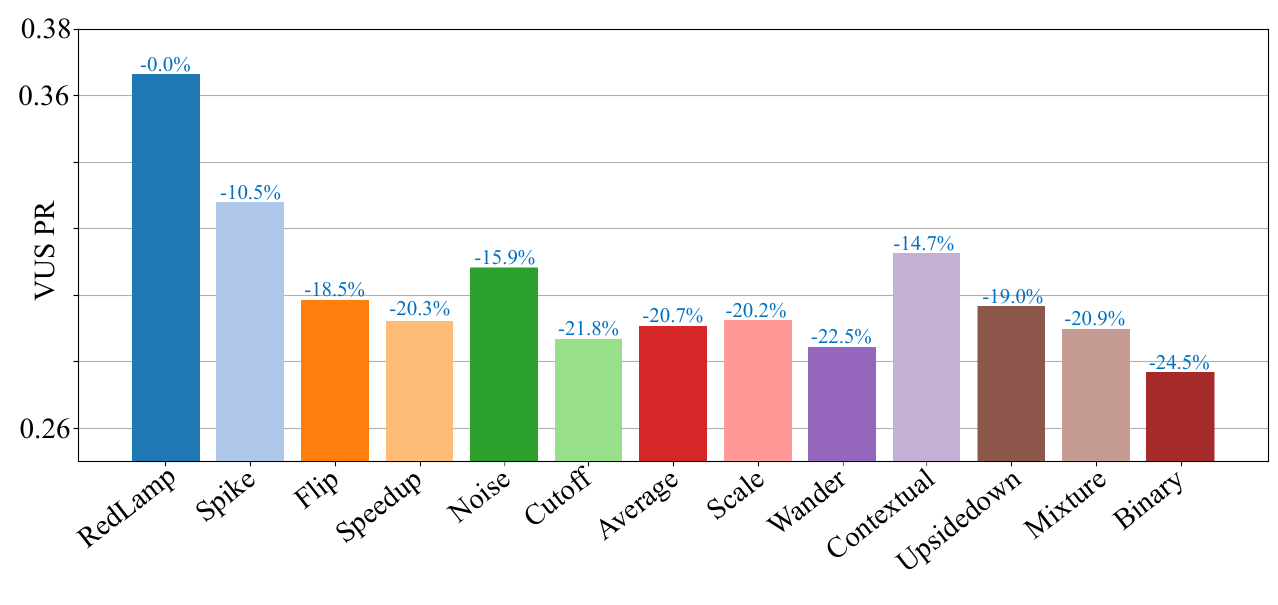} \\[-1em]
        (b) \smd \\
    \end{tabular}
    \vspace{-1.em}
    \caption{
        Effective data augmentations.
    }
    \label{fig:augfunc}
    \vspace{-1.em}
\end{figure}
We investigate the effectiveness of different data augmentations.
For this analysis, we perform binary classification for each data augmentation. Similar to the evaluation in \secton{\ref{sec:q2}}, Binary incorporates all data augmentations.
\fig{\ref{fig:augfunc}} presents the results for (a) \ucr and (b) \smd.
With \ucr, only Flip, Cutoff, and Contextual achieve higher accuracy compared to Binary.
Conversely, with \smd, Binary shows the lowest performance, which is attributed to an insufficient quantity of pseudo-anomalies.
When the augmentation ratio is increased to 4:1 w.r.t. Normal as in \fig{\ref{fig:quantity}} (b), VUS-PR performance improves to $0.312 (-14.8\%)$, surpassing many other data augmentations.
Spike yields the best results for \smd, probably because real anomalies in \smd often exhibit spike patterns.
These findings highlight the difficulty of manually designing data augmentation because the effective data augmentation varies depending on the dataset.
\method consistently outperforms all other settings, which indicates that our approach of utilizing all data augmentations in conjunction with multiclass classification is the optimal approach.

\begin{figure}[t]
    \centering
    \begin{tabular}{ccc}
        \includegraphics[width=0.3\linewidth]{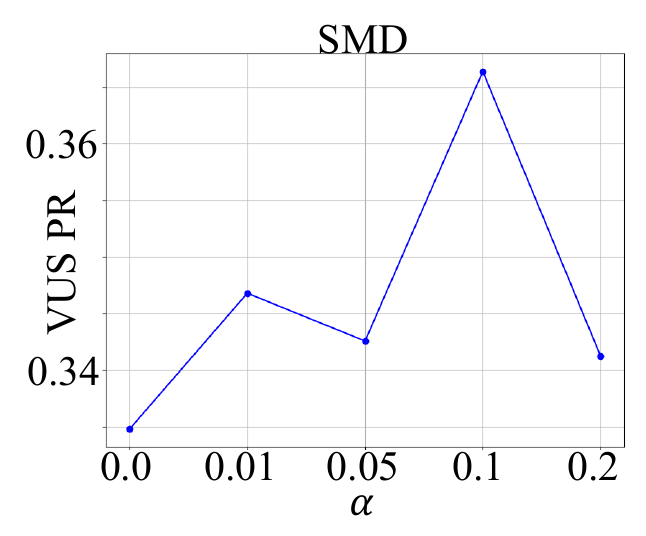} & 
        \includegraphics[width=0.3\linewidth]{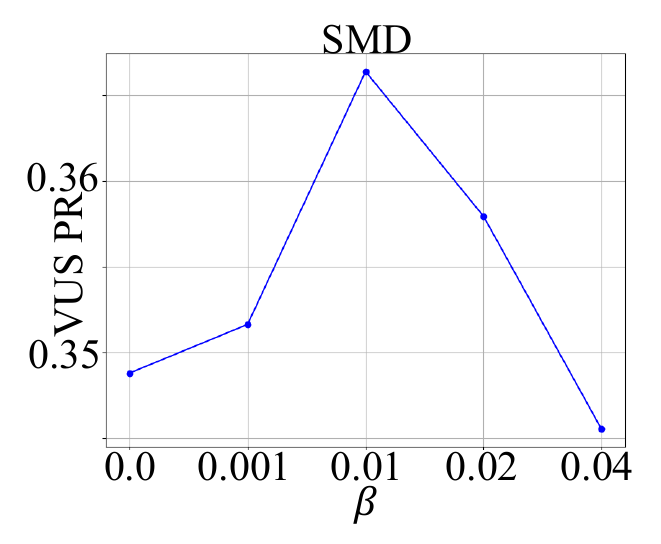} & 
        \includegraphics[width=0.3\linewidth]{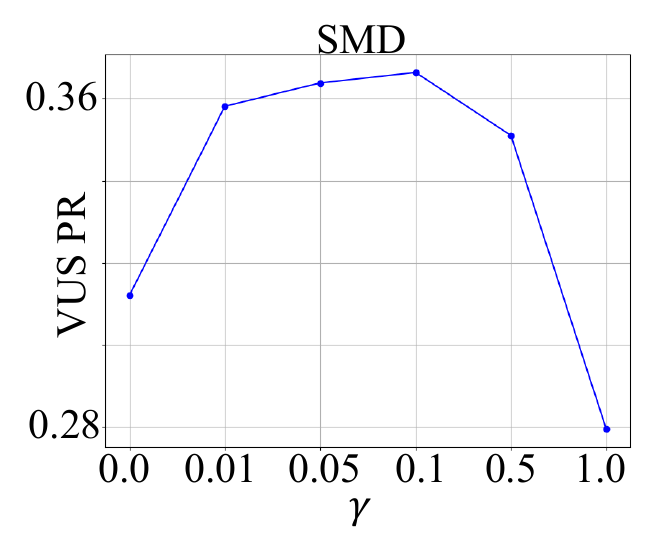} \\
        (a) $\probnorm$ & (b) $\probanom$ & (c) $\lossweight$ \\ 
    \end{tabular}
    \vspace{-1.em}
    \caption{
        Hyperparameter studies.
    }
    \label{fig:hyperparameter}
    \vspace{-1.em}
\end{figure}
\section{Hyperparamter Sensitivity} \label{apdx:hyperparameter}
\fig{\ref{fig:hyperparameter}} shows the effect of (a) $\probnorm$, (b) $\probanom$, and (c) $\lossweight$.

\myparaitemize{Effect of misclassified probability}
$\probnorm$ represents the probability that a pseudo-anomaly is classified as normal,
and $\probanom$ denotes the probability that one class is misclassified as another.
We vary $\probnorm$ across $\{0.0, \dots 0.2\}$, and $\probanom$ across $\{0.0, \dots 0.04\}$.
The results in \fig{\ref{fig:hyperparameter}} (a) and (b) show that setting both misclassified probabilities is essential and probabilities that are too large or small limit the performance.

\myparaitemize{Effect of loss weight}
\fig{\ref{fig:hyperparameter}} (c) shows the results of varying the loss weight $\lossweight$, which controls a balance between the reconstruction and cross-entropy loss.
We varied $\lossweight$ across $\{0.0, \dots 1.0\}$, where $\lossweight=0.0$ means the model is trained using solely $\lossfuncmse$ and $\lossweight=1.0$ means it is trained using solely $\lossfuncce$.
This emphasizes the importance of utilizing both loss functions to achieve optimal detection performance.

\end{document}